\documentclass[10pt,a4paper]{article}
\usepackage[utf8]{inputenc}
\usepackage{graphicx}
\usepackage{subcaption}
\usepackage[hyphens]{url}
\usepackage{hyperref}
\usepackage[round]{natbib}
\usepackage{booktabs}
\usepackage{makecell}
\usepackage{amsmath,amssymb}
\usepackage{geometry}
\graphicspath{{./figures/}{./}}

\title{Hybrid Topic-Semantic Labeling and Graph Embeddings for Unsupervised Legal Document Clustering}

\author{Deepak Bastola\footnote{Corresponding author: dxbastola@salisbury.edu} \\ Department of Mathematical Sciences \\ Salisbury University \\ dxbastola@salisbury.edu \and Woohyeok Choi \\ Department of Mathematics and Statistics \\ Carleton College \\ choir@carleton.edu}
\date{}

\begin{document}
\maketitle

\begin{abstract}
\noindent Legal documents pose substantial challenges for unsupervised knowledge discovery due to specialized domain terminology, multi-scale document lengths, and the scarcity of expert-annotated labels. This paper introduces and evaluates Topic-Graph Spherical Fusion (TGSF), a dual-manifold representation learning framework for unsupervised legal document clustering that unifies dense topic-semantic discovery and relational graph embeddings on a joint unit hypersphere. Instantiating dense topic discovery via Top2Vec and second-order biased random walks via Node2Vec, TGSF constructs a connected multi-relational document--topic graph $G = (V_D \cup V_T, E_{DT} \cup E_{DD}, W)$ that combines multi-topic document--centroid bridges with temperature-sharpened semantic $k$-Nearest Neighbor ($k$-NN) document--document edges ($\dim\ker(\mathcal{L}) = 1$). The resulting $L_2$-normalized semantic and graph-structural representations are fused via modality-weighted concatenation onto a joint unit hypersphere ($\mathbb{S}^{d_{\text{sem}}+d_{\text{graph}}-1}$) prior to KMeans partitioning, ensuring that squared chordal distance decomposes into a convex combination of semantic and structural cosine similarities. Evaluations across three benchmark legal corpora (the Atticus Clause Retrieval Dataset (ACORD), the Contract Understanding Atticus Dataset (CUAD), and the Labeled EDGAR (LEDGAR) provision corpus) show that TGSF consistently outperforms semantic-only ($\alpha=0$), structural-only ($\alpha=1$), Latent Dirichlet Allocation (LDA), Non-Negative Matrix Factorization (NMF), and Term Frequency--Inverse Document Frequency (TF-IDF) baselines. Across 216 factorial grid configurations and 240 Bayesian Tree-structured Parzen Estimator (TPE) and Multi-Objective NSGA-II Pareto optimization trials per dataset ($720$ trials total), TGSF improves Normalized Mutual Information (NMI) from $0.606 \to 0.653$ on LEDGAR (Adjusted Rand Index $\text{ARI} = 0.425$), $0.228 \to 0.284$ on ACORD ($\text{ARI} = 0.260$), and $0.535 \to 0.570$ on CUAD ($\text{ARI} = 0.282$). Functional Analysis of Variance (fANOVA) and Pareto frontier diagnostics further characterize hyperparameter sensitivity for reproducible legal text mining.

\vspace{8pt}
\noindent\textbf{Keywords:} Unsupervised Document Clustering $\cdot$ Topic-Graph Spherical Fusion $\cdot$ Graph Representation Learning $\cdot$ Dense Topic Discovery $\cdot$ Multi-Objective Bayesian Optimization $\cdot$ Legal Text Mining
\end{abstract}

\section{Introduction}
\label{sec:intro}

Legal texts are structurally complex, lengthy, and dense with domain-specific terminology and cross-referenced clauses, making manual annotation both costly and difficult to scale \citep{ariai2024legalnlpsurvey}. As digital repositories of commercial contracts, regulatory filings, and judicial opinions continue to expand, unsupervised document clustering has become an essential data mining primitive for legal information retrieval, contract due diligence, compliance auditing, and electronic discovery (e-discovery). When domain-specific training labels are scarce or unavailable, unsupervised representation learning enables practitioners to decompose large document collections into coherent functional groups and surface latent thematic structures automatically.

Early unsupervised approaches to legal text analysis relied primarily on lexical vector space models (such as TF-IDF paired with KMeans) or probabilistic and matrix-factorization topic models, notably Latent Dirichlet Allocation (LDA) \citep{blei2003lda} and Non-Negative Matrix Factorization (NMF) \citep{lee1999learning}, to discover latent themes in legislative and judicial collections \citep{didwania2024unveiling, sharaff2016email, oneill2016legislative}. While these techniques serve as foundational baselines in topic modeling, they require pre-specifying the number of topics and ignore word order and contextual semantics, making them sensitive to document length, boilerplate phrasing, and jurisdiction-specific vocabulary. Conversely, pretrained transformer encoders such as Sentence-BERT \citep{reimers2019sentencebert} and domain-adapted Legal-BERT \citep{chalkidis2020legalbert, chalkidis2019eurlex, zheng2021when} capture rich contextual semantics, and neural topic frameworks such as Top2Vec \citep{angelov2020top2vec} automatically identify dense semantic topic centroids in a joint document--word space without stop-word removal or stemming. However, purely semantic embeddings treat documents as isolated points in vector space, neglecting higher-order relational transitivity across documents that share secondary thematic affinities or local neighborhood bridges.

Graph representation learning offers a complementary perspective by encoding multi-hop relational structure among documents and topics \citep{grover2016node2vec, lodha2019legalnode2vec, bhattacharya2020similarity, bhattacharya2022legal, liu2026deepgraph}. Yet a fundamental topological challenge arises when constructing document--topic networks in an unsupervised setting: if each document node is linked exclusively to its single highest-scoring topic centroid without inter-document links, the resulting bipartite graph decomposes into $T$ mutually disconnected star subgraphs whose normalized graph Laplacian exhibits a $T$-dimensional nullspace ($\dim\ker(\mathcal{L}_{\text{bipartite}}) = T$). Random walks trapped within disconnected star components collapse intra-topic variance without capturing inter-topic proximity. Furthermore, naively concatenating unnormalized semantic and graph embeddings allows whichever modality has larger coordinate variance to dominate Euclidean clustering objectives.

\begin{figure}[!htbp]
  \centering
  \includegraphics[width=0.94\textwidth]{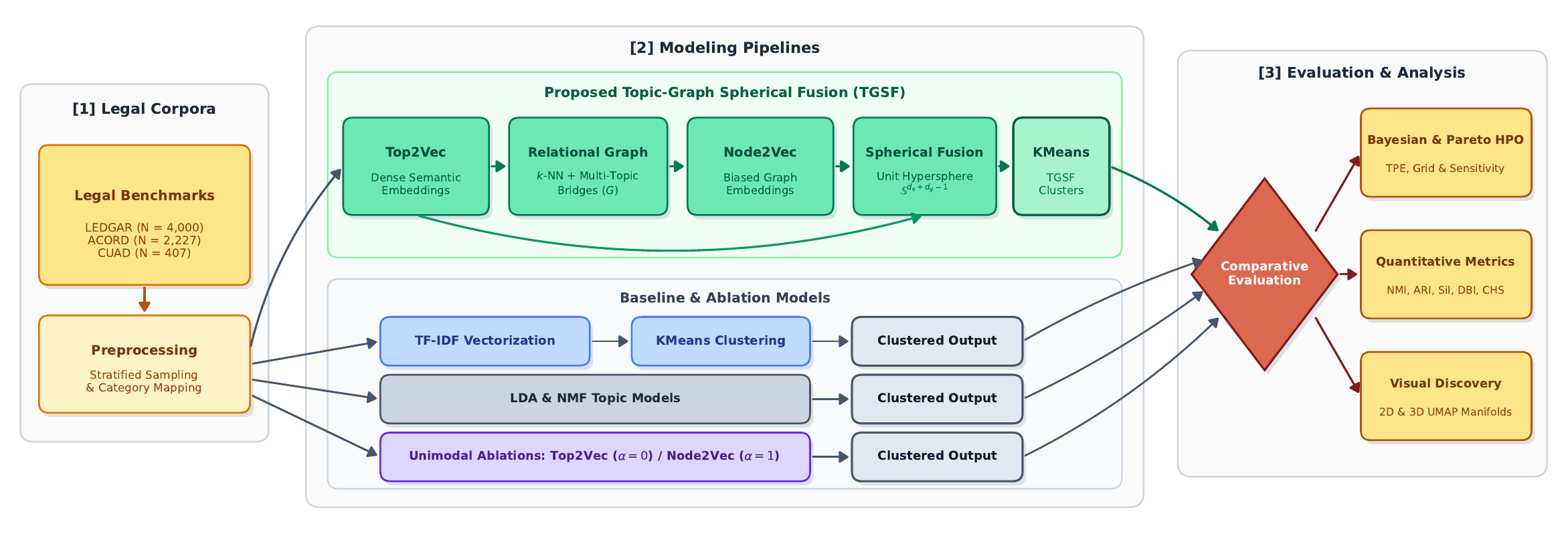}
  \caption{Architectural overview of the proposed Topic-Graph Spherical Fusion (TGSF) framework (integrating Top2Vec, connected multi-relational graph $G$, second-order Node2Vec, and unit-hypersphere $L_2$ fusion) alongside lexical (TF-IDF + KMeans), probabilistic/matrix-factorization (LDA, NMF), and unimodal semantic ($\alpha=0$) and structural ($\alpha=1$) baselines across LEDGAR, ACORD, and CUAD.}
  \label{fig:pipeline}
\end{figure}

To address these methodological challenges, we propose Topic-Graph Spherical Fusion (TGSF), a unified unsupervised dual-manifold representation framework that couples dense semantic topic discovery, multi-relational graph construction, second-order biased random walks, and spherical modality fusion. As illustrated in Figure~\ref{fig:pipeline}, TGSF first maps documents and automatically discovered topic centroids into a shared semantic space via Top2Vec \citep{angelov2020top2vec}. Second, it constructs a connected multi-relational graph $G = (V_D \cup V_T, E_{DT} \cup E_{DD}, W)$ that bridges topic stars through top-$M$ multi-topic document--centroid affinities ($E_{DT}$) and temperature-sharpened semantic $k$-Nearest Neighbor ($k$-NN) document--document edges ($E_{DD}$), guaranteeing a single connected component ($\dim\ker(\mathcal{L}) = 1$). Third, TGSF learns structural embeddings via biased second-order Node2Vec walks \citep{grover2016node2vec} and fuses the $L_2$-normalized semantic and structural vectors onto the joint unit hypersphere $\mathbb{S}^{d_{\text{sem}}+d_{\text{graph}}-1}$ prior to KMeans partitioning \citep{macqueen1967kmeans} and Uniform Manifold Approximation and Projection (UMAP) visualization \citep{mcinnes2018umap}.

\paragraph{Summary of Contributions.} The primary contributions of this work are fourfold:
\begin{enumerate}
  \item \textbf{Multi-Relational Topic--Document Graph and Spectral Connectivity:} We formulate a weighted multi-relational graph $G = (V_D \cup V_T, E_{DT} \cup E_{DD}, W)$ combining thresholded multi-topic document--centroid bridges ($E_{DT}$) with power-sharpened semantic $k$-NN document--document edges ($E_{DD}$), guaranteeing a connected manifold ($\dim\ker(\mathcal{L}) = 1$) that supports ergodic cross-cluster random walks.
  \item \textbf{Spherical Modality Fusion on $\mathbb{S}^{d_{\text{sem}}+d_{\text{graph}}-1}$:} We introduce a modality-weighted spherical fusion operator $\mathbf{h}_i(\alpha) = [\sqrt{1-\alpha}\,\hat{\mathbf{e}}_i \,\|\, \sqrt{\alpha}\,\hat{\mathbf{z}}_i] \in \mathbb{S}^{d_{\text{sem}}+d_{\text{graph}}-1}$ whose squared chordal distance decomposes into a convex kernel combination of semantic and structural cosine similarities, preventing scale or variance dominance by either modality during KMeans clustering.
  \item \textbf{Multi-Dataset Legal Benchmarking and Bayesian/Pareto Optimization:} We evaluate TGSF across three legal corpora spanning distinct structural granularities: LEDGAR ($N=4,000$ functional contract provisions) \citep{tuggener2020ledgar}, ACORD ($N=2,227$ expert-annotated retrieval clauses) \citep{wang2025acord}, and CUAD ($N=407$ full SEC commercial agreements) \citep{hendrycks2021cuad}, conducting 216 factorial grid evaluations and 240 GPU-accelerated Bayesian Multivariate TPE and Multi-Objective NSGA-II Pareto trials per dataset ($720$ HPO trials in total).
  \item \textbf{Variance Decomposition and Reproducible Pipeline:} Using functional ANOVA (fANOVA), we quantify the relative contribution of semantic encoder backbones (Legal-BERT, MPNet, MiniLM), graph topology parameters ($\tau, \beta, \gamma, k_{\text{NN}}, M$), random-walk biases ($p, q$), and fusion weight ($\alpha$), and release a complete CPU/GPU implementation for reproducible legal knowledge discovery.
\end{enumerate}

The remainder of this paper is organized as follows. Section~\ref{sec:related_work} reviews related work in unsupervised topic modeling, graph representation learning, and multi-modal legal document clustering. Section~\ref{sec:methodology} formulates the mathematical architecture of TGSF. Section~\ref{sec:setup} details the benchmark legal corpora, baseline models, validation metrics, and hyperparameter search spaces. Section~\ref{sec:results} presents the empirical results, UMAP manifold visualizations, factorial grid sensitivity, Bayesian/Pareto optimization studies, and ablations. Section~\ref{sec:discussion} discusses domain insights, computational scalability, and practical deployment guidelines, and Section~\ref{sec:conclusion} concludes the paper.

\section{Related Work}
\label{sec:related_work}

\subsection{Unsupervised Topic Modeling and Dense Semantic Clustering}
\label{subsec:related_topic}
Classical topic models represent documents as mixtures over latent word distributions. Latent Dirichlet Allocation (LDA) \citep{blei2003lda} models document--topic and topic--word distributions via Dirichlet priors, whereas Non-Negative Matrix Factorization (NMF) \citep{lee1999learning} decomposes term--document matrices into additive non-negative factors. Both methods have been widely applied to legislative acts \citep{oneill2016legislative}, judicial proceedings \citep{didwania2024unveiling}, and e-discovery collections \citep{sharaff2016email}. However, bag-of-words representations suffer from lexical sparsity and cannot distinguish polysemous legal terms or synonymous contractual phrasing.

Recent neural topic discovery methods overcome lexical sparsity by projecting documents into continuous embedding spaces learned by pretrained transformers \citep{reimers2019sentencebert, chalkidis2020legalbert}. Top2Vec \citep{angelov2020top2vec} applies UMAP dimensionality reduction \citep{mcinnes2018umap} followed by HDBSCAN density clustering \citep{campello2013hdbscan} to identify dense semantic regions and extract topic centroids automatically without requiring a preset topic count or stop-word removal. While dense semantic embeddings capture local contextual similarity, they do not explicitly encode multi-hop structural relationships across documents that bridge multiple legal themes.

\subsection{Graph Representation Learning and Document Networks}
\label{subsec:related_graph}
Network-based document analysis models corpora as graphs in which nodes represent documents, citations, or thematic entities. Node2Vec \citep{grover2016node2vec} learns low-dimensional node representations by optimizing a Skip-gram objective over truncated random walks, introducing second-order return ($p$) and in-out ($q$) bias parameters to interpolate between breadth-first community homophily and depth-first structural equivalence. In the legal domain, \citet{lodha2019legalnode2vec}, \citet{bhattacharya2020similarity}, and \citet{bhattacharya2022legal} demonstrated that citation and precedent graphs embedded via Node2Vec improve statute and case-law similarity retrieval, while \citet{liu2026deepgraph} surveyed deep graph clustering on relational networks. Unlike citation networks (which are unavailable for standalone commercial contracts or unlinked contract clauses), our TGSF framework constructs a self-contained, connected multi-relational graph directly from discovered topic centroids and semantic $k$-NN neighborhoods.

\subsection{Multi-Modal Fusion and Legal Knowledge Discovery}
\label{subsec:related_fusion}
Combining textual semantics with structural graph signals has been shown to improve representation quality in legal document analysis \citep{bhattacharya2022legal, liu2026deepgraph}. However, existing legal NLP studies largely focus on supervised classification over labeled benchmarks such as EURLEX \citep{chalkidis2019eurlex}, LexGLUE/LEDGAR \citep{tuggener2020ledgar}, CUAD \citep{hendrycks2021cuad}, and ACORD \citep{wang2025acord}, where models are trained via cross-entropy loss on thousands of expert annotations \citep{zheng2021when, ariai2024legalnlpsurvey}. In contrast, TGSF operates in a strictly unsupervised regime: it uses ground-truth labels solely for post-hoc external validation (NMI and ARI), while introducing spherical modality fusion on $\mathbb{S}^{d_{\text{sem}}+d_{\text{graph}}-1}$ and multi-objective Bayesian Pareto optimization to reconcile internal geometric compactness with external category recovery.

\section{Methodology: Topic-Graph Spherical Fusion (TGSF)}
\label{sec:methodology}

Let $\mathcal{D} = \{d_1, d_2, \dots, d_N\}$ denote an unlabeled corpus of $N$ legal documents. The Topic-Graph Spherical Fusion (TGSF) architecture integrates continuous semantic embeddings and relational graph representation learning in a four-stage framework to learn a unified, scale-invariant representation $\mathbf{h}_i \in \mathbb{S}^{d_{\text{sem}}+d_{\text{graph}}-1}$ for each document $d_i$ and partition $\mathcal{D}$ into $K$ coherent legal clusters $\mathcal{C} = \{\mathcal{C}_1, \dots, \mathcal{C}_K\}$.

\subsection{Dense Semantic Topic Discovery via Top2Vec}
\label{subsec:top2vec}
To extract continuous document representations and latent thematic anchors without manual supervision, we first apply Top2Vec \citep{angelov2020top2vec} utilizing a pretrained transformer encoder $f_{\theta} : \mathcal{D} \to \mathbb{R}^{d_{\text{sem}}}$ (evaluating all-MiniLM-L6-v2 \citep{reimers2019sentencebert} with $d_{\text{sem}} = 384$, all-mpnet-base-v2 with $d_{\text{sem}} = 768$, and domain-pretrained Legal-BERT \citep{chalkidis2020legalbert} with $d_{\text{sem}} = 768$) to map each document $d_i \in \mathcal{D}$ to a dense semantic vector $\mathbf{e}_i = f_{\theta}(d_i) \in \mathbb{R}^{d_{\text{sem}}}$.

To discover latent thematic structure without pre-specifying the number of topics, Top2Vec projects $\{\mathbf{e}_i\}_{i=1}^N$ into a low-dimensional manifold via UMAP \citep{mcinnes2018umap} and identifies $T$ dense document clusters $\{\mathcal{C}_1^{\text{top}}, \dots, \mathcal{C}_T^{\text{top}}\}$ via HDBSCAN \citep{campello2013hdbscan}. Each discovered topic $t_m \in V_T$ ($m \in \{1, \dots, T\}$) is parameterized by its semantic centroid $\boldsymbol{\mu}_m \in \mathbb{R}^{d_{\text{sem}}}$, inducing a continuous document--topic cosine affinity matrix $S^{(DT)} \in [-1, 1]^{N \times T}$:
\begin{equation}
  \boldsymbol{\mu}_m = \frac{1}{|\mathcal{C}_m^{\text{top}}|} \sum_{d_i \in \mathcal{C}_m^{\text{top}}} \mathbf{e}_i, \qquad
  s_{i,m} = \cos(\mathbf{e}_i, \boldsymbol{\mu}_m) = \frac{\mathbf{e}_i^\top \boldsymbol{\mu}_m}{\|\mathbf{e}_i\|_2 \, \|\boldsymbol{\mu}_m\|_2}.
  \label{eq:top2vec_affinity}
\end{equation}
Similarly, pairwise semantic similarity between any two documents $d_i, d_j \in \mathcal{D}$ ($i \neq j$) is given by $S_{ij} = \cos(\mathbf{e}_i, \mathbf{e}_j) = \mathbf{e}_i^\top \mathbf{e}_j / (\|\mathbf{e}_i\|_2 \|\mathbf{e}_j\|_2)$.

\subsection{Multi-Relational Graph Construction and Spectral Connectivity}
\label{subsec:graph_construction}

Subsequent to topic discovery, we construct an undirected weighted relational graph $G = (V, E, W)$ over the vertex set $V = V_D \cup V_T$, where $V_D = \{d_1, \dots, d_N\}$ denotes the $N$ document nodes and $V_T = \{t_1, \dots, t_T\}$ denotes the $T$ topic hub nodes ($|V| = N + T$).

\begin{figure}[!htbp]
  \centering
  \includegraphics[width=0.86\textwidth]{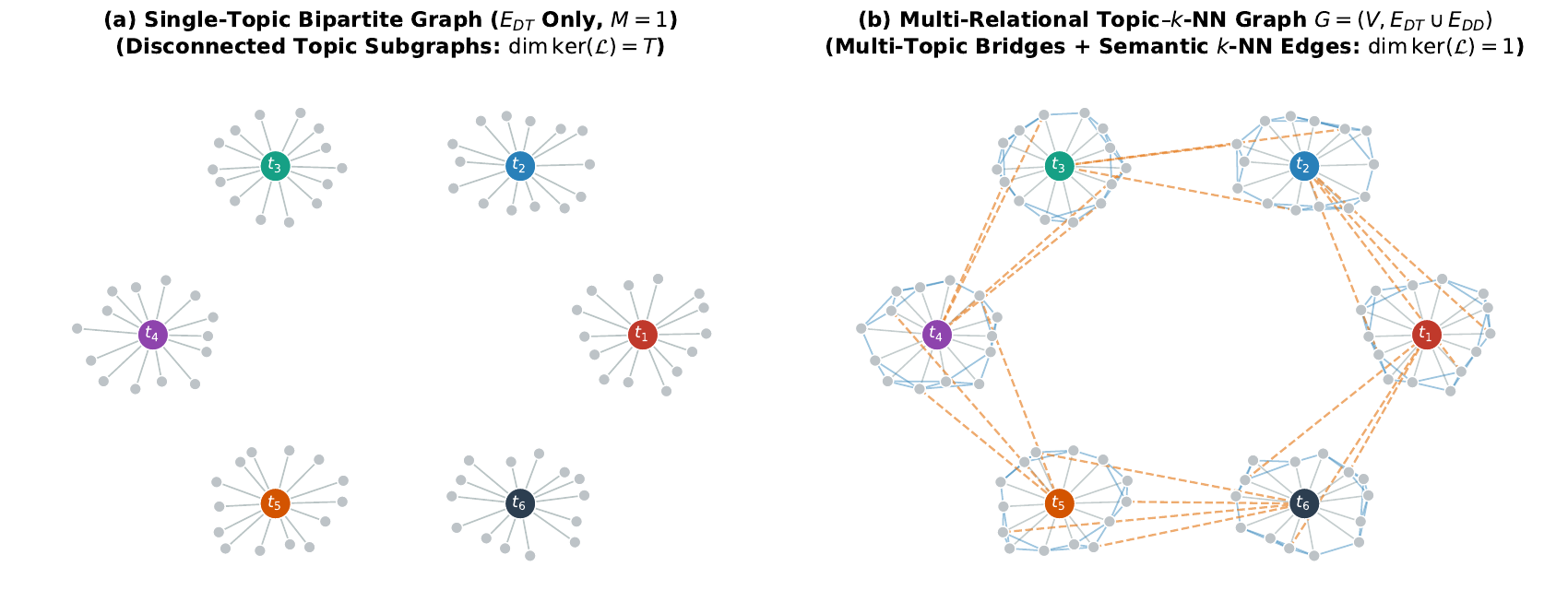}
  \caption{Topological and spectral comparison between (a) a single-topic bipartite graph ($E_{DT}$ only, $M=1$), which decomposes into $T$ disconnected star subgraphs ($\dim\ker(\mathcal{L}) = T$), and (b) the TGSF multi-relational graph $G = (V_D \cup V_T, E_{DT} \cup E_{DD}, W)$, which integrates multi-topic bridges (dashed orange) and cosine-weighted semantic $k$-NN document--document edges (solid blue) to form a connected manifold ($\dim\ker(\mathcal{L}) = 1$).}
  \label{fig:bipartite}
\end{figure}

If a single-topic bipartite graph $G_{\text{bipartite}}$ connects each document $d_i \in V_D$ solely to its single highest-scoring topic node $t_{m^*(i)}$ where $m^*(i) = \arg\max_{1 \le m \le T} s_{i,m}$, its adjacency matrix $A_{\text{bipartite}}$ and normalized graph Laplacian $\mathcal{L}_{\text{bipartite}} = I - D_{\text{bipartite}}^{-1/2} A_{\text{bipartite}} D_{\text{bipartite}}^{-1/2}$ satisfy:
\begin{equation}
  A_{\text{bipartite}}(d_i, t_m) = s_{i,m} \, \mathbb{I}\!\left(m = m^*(i)\right), \quad
  G_{\text{bipartite}} = \bigsqcup_{m=1}^{T} S_m, \quad
  \dim \ker(\mathcal{L}_{\text{bipartite}}) = T > 1.
  \label{eq:naive_laplacian}
\end{equation}
Because every document node in $G_{\text{bipartite}}$ has degree $1$, the graph decomposes into $T$ mutually disconnected star components $\{S_1, \dots, S_T\}$, and by spectral graph theory the multiplicity of the $0$-eigenvalue of $\mathcal{L}_{\text{bipartite}}$ equals $T$ (Figure~\ref{fig:bipartite}a). Any Markovian random walk initialized in component $S_m$ has zero transition probability of reaching $S_{m'}$ ($m' \neq m$), preventing cross-topic structural propagation.

To guarantee global connectivity ($\dim\ker(\mathcal{L}) = 1$) while preserving community-anchoring topic hubs, TGSF constructs the multi-relational graph $G = (V_D \cup V_T, E_{DT} \cup E_{DD}, W)$, illustrated in Figure~\ref{fig:bipartite}(b). The edge weight matrix $W$ is defined over two complementary edge sets:
\begin{enumerate}
  \item \textbf{Multi-Topic Document--Topic Bridges ($E_{DT}$):} Let $\mathcal{T}_M(d_i) \subset V_T$ denote the top-$M$ topic hubs ranked by affinity $s_{i,m}$ for document $d_i$. An undirected edge $(d_i, t_m) \in E_{DT}$ is formed with scaled weight:
  \begin{equation}
    w_{DT}(d_i, t_m) = \gamma \, s_{i,m} \cdot \mathbb{I}\!\left(t_m \in \mathcal{T}_M(d_i) \;\wedge\; s_{i,m} > \tau\right),
    \label{eq:edt_weights}
  \end{equation}
  where $M \in \{1, \dots, 5\}$ controls multi-topic bridge multiplicity, $\tau > 0$ is a minimum relevance threshold, and $\gamma > 0$ scales document--topic hub attraction relative to document--document edges. Multi-topic bridges allow documents addressing overlapping legal concepts (e.g., an indemnification provision that also specifies governing law) to act as structural conduits between topic hubs.
  \item \textbf{Temperature-Sharpened Semantic $k$-NN Edges ($E_{DD}$):} Let $\mathcal{N}_{k_{\text{NN}}}(d_i)$ denote the $k_{\text{NN}}$ nearest semantic neighbors of $d_i$ under cosine similarity $S_{ij}$. An undirected edge $(d_i, d_j) \in E_{DD}$ is added with power-sharpened cosine weight:
  \begin{equation}
    w_{DD}(d_i, d_j) = \big(\max(0, S_{ij})\big)^{\beta} \cdot \mathbb{I}\!\left(d_j \in \mathcal{N}_{k_{\text{NN}}}(d_i) \;\vee\; d_i \in \mathcal{N}_{k_{\text{NN}}}(d_j)\right),
    \label{eq:edd_weights}
  \end{equation}
  where $\beta > 0$ acts as an inverse-temperature exponent sharpening high-confidence semantic links.
\end{enumerate}

The union of multi-topic bridges $E_{DT}$ and mutual $k$-NN edges $E_{DD}$ links the $T$ topic neighborhoods into a single connected graph component, reducing the nullspace of the normalized Laplacian $\mathcal{L}$ of $G$ to one dimension:
\begin{equation}
  \dim \ker(\mathcal{L}) = 1, \qquad 0 = \lambda_1(\mathcal{L}) < \lambda_2(\mathcal{L}) \le \dots \le \lambda_{N+T}(\mathcal{L}),
  \label{eq:laplacian_connected}
\end{equation}
where the strictly positive algebraic connectivity (Fiedler value $\lambda_2(\mathcal{L}) > 0$) ensures ergodic random-walk mixing across the entire document--topic manifold.

\subsection{Second-Order Biased Random-Walk Embeddings via Node2Vec}
\label{subsec:node2vec}

Given the weighted relational graph $G$, we apply Node2Vec \citep{grover2016node2vec} to learn structural representations $\mathbf{z}_u \in \mathbb{R}^{d_{\text{graph}}}$ for all nodes $u \in V$. Node2Vec simulates $R$ second-order biased random walks of length $L$ starting from each node. Having traversed edge $(u, v)$ to reside at node $v$ at step $r-1$, the transition probability to neighbor $x \in \mathcal{N}_G(v)$ at step $r$ is governed by the second-order Markov kernel:
\begin{equation}
  P(c_r = x \mid c_{r-1} = v,\, c_{r-2} = u) = \frac{\alpha_{pq}(u, x) \, w(v, x)}{\sum_{y \in \mathcal{N}_G(v)} \alpha_{pq}(u, y) \, w(v, y)}, \;
  \alpha_{pq}(u, x) = \begin{cases}
    1/p & \text{if } d_{G}(u, x) = 0, \\
    1   & \text{if } d_{G}(u, x) = 1, \\
    1/q & \text{if } d_{G}(u, x) = 2,
  \end{cases}
  \label{eq:node2vec_walk}
\end{equation}
where $d_G(u, x) \in \{0, 1, 2\}$ denotes the shortest-path topological distance between predecessor $u$ and candidate $x$, $p > 0$ is the return parameter controlling local revisit probability, and $q > 0$ is the in-out parameter balancing breadth-first community homophily ($q > 1$) against depth-first structural equivalence ($q < 1$).

The node embeddings $\mathbf{Z} = \{\mathbf{z}_u\}_{u \in V}$ are optimized via stochastic gradient ascent on the Skip-gram negative-sampling log-likelihood over walk context windows $\mathcal{N}_S(u)$ of radius $w$:
\begin{equation}
  \max_{\mathbf{Z}} \; \mathcal{L}_{\text{N2V}}(\mathbf{Z}) = \sum_{u \in V} \sum_{v \in \mathcal{N}_S(u)} \left[ \log \sigma\!\left(\mathbf{z}_v^\top \mathbf{z}_u\right) + \sum_{k=1}^{K_{\text{neg}}} \mathbb{E}_{v_k \sim P_n(V)}\!\left[ \log \sigma\!\left(-\mathbf{z}_{v_k}^\top \mathbf{z}_u\right) \right] \right],
  \label{eq:node2vec_loss}
\end{equation}
where $\sigma(x) = (1 + e^{-x})^{-1}$ is the logistic sigmoid function and $P_n(V) \propto \text{deg}(v)^{3/4}$ is the unigram negative-sampling noise distribution. Because random walks on $G$ traverse both shared topic hubs ($d_i \to t_m \to d_j$) and direct $k$-NN links ($d_i \to d_j$), Equation~\eqref{eq:node2vec_loss} embeds documents that share multi-hop topical communities and local semantic neighborhoods into proximate regions of $\mathbb{R}^{d_{\text{graph}}}$. Restricting $\mathbf{Z}$ to $V_D$ yields the document structural representations $\{\mathbf{z}_1, \dots, \mathbf{z}_N\} \subset \mathbb{R}^{d_{\text{graph}}}$.

\subsection{Spherical Modality Fusion on \texorpdfstring{$\mathbb{S}^{d_{\text{sem}}+d_{\text{graph}}-1}$}{S(d\_sem+d\_graph-1)}, KMeans, and UMAP}
\label{subsec:spherical_fusion}

Directly concatenating raw semantic vectors $\mathbf{e}_i \in \mathbb{R}^{d_{\text{sem}}}$ and graph vectors $\mathbf{z}_i \in \mathbb{R}^{d_{\text{graph}}}$ introduces severe scale imbalance whenever $\|\mathbf{e}_i\|_2 \neq \|\mathbf{z}_i\|_2$. To combine both modalities onto a common geometric manifold without variance or scale dominance, TGSF projects each modality onto its respective unit hypersphere ($\hat{\mathbf{e}}_i \in \mathbb{S}^{d_{\text{sem}}-1}$ and $\hat{\mathbf{z}}_i \in \mathbb{S}^{d_{\text{graph}}-1}$) and maps each document onto the joint unit hypersphere $\mathbb{S}^{d_{\text{sem}} + d_{\text{graph}} - 1}$ controlled by modality fusion weight $\alpha \in [0, 1]$:
\begin{equation}
  \hat{\mathbf{e}}_i = \frac{\mathbf{e}_i}{\|\mathbf{e}_i\|_2}, \qquad
  \hat{\mathbf{z}}_i = \frac{\mathbf{z}_i}{\|\mathbf{z}_i\|_2}, \qquad
  \mathbf{h}_i(\alpha) = \left[ \sqrt{1 - \alpha}\,\hat{\mathbf{e}}_i \;\Big\|\; \sqrt{\alpha}\,\hat{\mathbf{z}}_i \right] \in \mathbb{S}^{d_{\text{sem}} + d_{\text{graph}} - 1}.
  \label{eq:concat}
\end{equation}

\paragraph{Chordal Distance Decomposition Property.} Because $\|\mathbf{h}_i(\alpha)\|_2^2 = (1-\alpha)\|\hat{\mathbf{e}}_i\|_2^2 + \alpha\|\hat{\mathbf{z}}_i\|_2^2 = 1$ for all $\alpha \in [0, 1]$, the squared Euclidean (chordal) distance between any two fused document representations on $\mathbb{S}^{d_{\text{sem}}+d_{\text{graph}}-1}$ decomposes into a convex kernel combination of semantic and structural cosine similarities:
\begin{equation}
  \big\|\mathbf{h}_i(\alpha) - \mathbf{h}_j(\alpha)\big\|_2^2 = 2 - 2\Big[(1 - \alpha)\cos(\mathbf{e}_i, \mathbf{e}_j) + \alpha\cos(\mathbf{z}_i, \mathbf{z}_j)\Big].
  \label{eq:chordal_decomposition}
\end{equation}
Equation~\eqref{eq:chordal_decomposition} guarantees that $\alpha = 0$ recovers pure semantic cosine clustering (Top2Vec-only), $\alpha = 1$ recovers pure structural graph clustering (Node2Vec-only), and $\alpha = 0.5$ assigns equal unit norm to both modalities.

We then partition the fused spherical representations $\{\mathbf{h}_1(\alpha), \dots, \mathbf{h}_N(\alpha)\}$ into $K$ disjoint clusters $\mathcal{C} = \{\mathcal{C}_1, \dots, \mathcal{C}_K\}$ with centroids $\{\mathbf{c}_1, \dots, \mathbf{c}_K\}$ by minimizing the KMeans within-cluster sum-of-squares objective \citep{macqueen1967kmeans} with k-means++ initialization:
\begin{equation}
  \min_{\{\mathcal{C}_k\}_{k=1}^K,\, \{\mathbf{c}_k\}_{k=1}^K} \mathcal{J}_{\text{KMeans}}(\mathcal{C}; \alpha) = \sum_{k=1}^{K} \sum_{d_i \in \mathcal{C}_k} \big\| \mathbf{h}_i(\alpha) - \mathbf{c}_k \big\|_2^2, \qquad
  \mathbf{c}_k = \frac{1}{|\mathcal{C}_k|} \sum_{d_i \in \mathcal{C}_k} \mathbf{h}_i(\alpha).
  \label{eq:kmeans_obj}
\end{equation}
For low-dimensional visual inspection, we project $\{\mathbf{h}_i(\alpha)\}_{i=1}^N$ into $\mathbb{R}^2$ and $\mathbb{R}^3$ using UMAP \citep{mcinnes2018umap}, which models the high-dimensional data as a fuzzy simplicial set and minimizes fuzzy set cross-entropy (see Appendix~\ref{sec:UMAP-theory}, Equations~\eqref{eq:umap_pji}--\eqref{eq:umap_loss} for the complete mathematical derivation).

\section{Experimental Setup}
\label{sec:setup}

\subsection{Benchmark Legal Corpora}
\label{subsec:datasets}
To evaluate the generalizability of TGSF across legal text granularities (ranging from short retrieval clauses and functional provisions to full multi-page commercial contracts), we conduct experiments on three benchmark datasets summarized in Table~\ref{tab:dataset_summary}:
\begin{enumerate}
  \item \textbf{LEDGAR (Labeled EDGAR Provisions)} \citep{tuggener2020ledgar}: A large-scale corpus of corporate contract provisions extracted from U.S.\ Securities and Exchange Commission (SEC) EDGAR filings (LexGLUE benchmark). We evaluate a stratified sample of $N=4,000$ provisions spanning $K=28$ primary functional categories (e.g., Governing Law, Indemnification, Termination, Notices, Severability, and Waivers), where Top2Vec discovers $T=73$ fine-grained semantic topics.
  \item \textbf{ACORD (Atticus Clause Retrieval Dataset)} \citep{wang2025acord}: An expert-annotated clause retrieval benchmark for contract drafting containing $N=2,227$ labeled clause--query pairs spanning $K=7$ macro-level legal categories and $25$ fine-grained clause types, with $T=25$ discovered Top2Vec topics.
  \item \textbf{CUAD (Contract Understanding Atticus Dataset)} \citep{hendrycks2021cuad}: A benchmark of full-length commercial agreements extracted from SEC EDGAR filings. Deduplicating contract contexts yields $N=407$ distinct multi-page agreements across $24$ official SEC agreement categories (e.g., Strategic Alliance, License, Distributor, Maintenance, Sponsorship, Development, Service, Hosting, Outsourcing, Co-Branding, Franchise, and Joint Venture), evaluated at $K=20$ (with $T=10$ discovered macro-topics) and across $K \in [5, 30]$ in sensitivity analysis.
\end{enumerate}

\begin{table}[!htbp]
  \centering
  \caption{Summary Characteristics of the Three Legal Benchmark Corpora.}
  \label{tab:dataset_summary}
  \resizebox{\linewidth}{!}{%
  \begin{tabular}{@{}llrrrr@{}}
    \toprule
    \textbf{Benchmark} & \textbf{Legal Text Granularity} & \textbf{Documents ($N$)} & \textbf{Discovered Topics ($T$)} & \textbf{Eval.\ Clusters ($K$)} & \textbf{Source Repository} \\
    \midrule
    LEDGAR \citep{tuggener2020ledgar} & Functional Contract Provisions & $4,000$ & $73$ & $28$ & SEC EDGAR / LexGLUE \\
    ACORD \citep{wang2025acord}       & Retrieval Clauses \& Queries   & $2,227$ & $25$ & $7$  & Atticus Project \\
    CUAD \citep{hendrycks2021cuad}    & Full Commercial Agreements     & $407$   & $10$ & $20$ & SEC EDGAR / Atticus \\
    \bottomrule
  \end{tabular}%
  }
\end{table}

\begin{figure}[!htbp]
  \centering
  \includegraphics[width=0.92\textwidth]{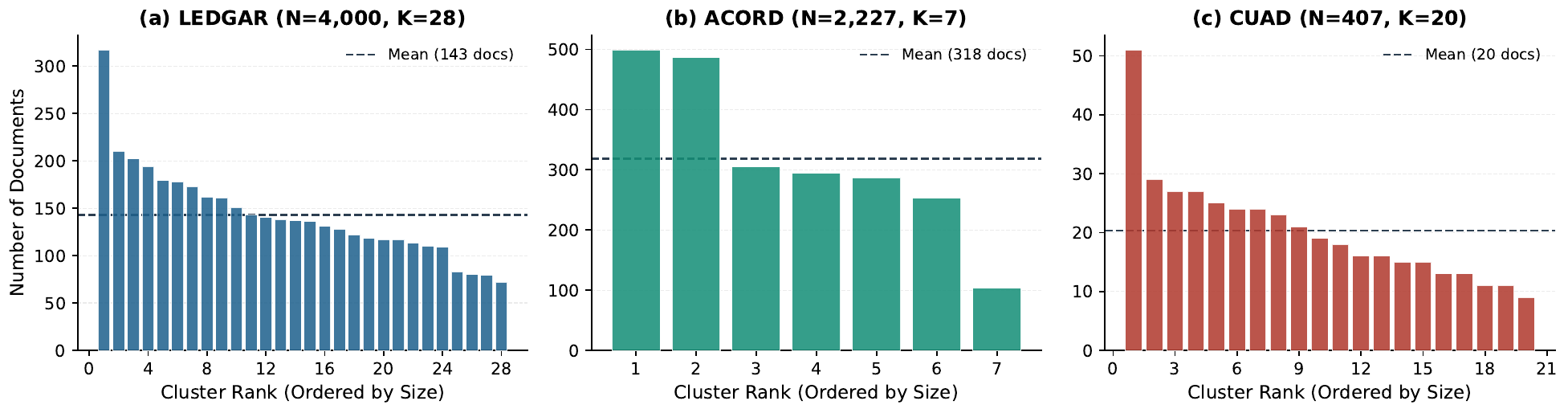}
  \caption{Cluster size distributions produced by TGSF across (a) LEDGAR ($N=4,000, K=28$), (b) ACORD ($N=2,227, K=7$), and (c) CUAD ($N=407, K=20$), ordered by cluster rank with dashed horizontal lines indicating the mean cluster size.}
  \label{fig:barchart}
\end{figure}

Figure~\ref{fig:barchart} illustrates the cluster size distributions produced by TGSF across all three benchmarks, confirming that the framework produces balanced, well-populated partitions reflective of the underlying thematic distributions without degenerate singleton clusters.

\subsection{Baseline Models and Unimodal Ablations}
\label{subsec:baselines}
We compare TGSF against five unsupervised baselines and unimodal ablations:
\begin{enumerate}
  \item \textbf{TF-IDF + KMeans:} $L_2$-normalized term-frequency inverse-document-frequency vectors ($5,000$ maximum features, English stop words removed) partitioned via KMeans.
  \item \textbf{Latent Dirichlet Allocation (LDA)} \citep{blei2003lda}: Variational Bayes document--topic simplex distributions partitioned by dominant topic assignment.
  \item \textbf{Non-Negative Matrix Factorization (NMF)} \citep{lee1999learning}: Frobenius-norm TF-IDF factorization initialized via NNDSVDA.
  \item \textbf{Top2Vec Only (Semantic Ablation, $\alpha = 0$):} Unit-normalized dense semantic embeddings $\hat{\mathbf{e}}_i \in \mathbb{S}^{d_{\text{sem}}-1}$ clustered directly via KMeans without graph structure.
  \item \textbf{Node2Vec Only (Structural Ablation, $\alpha = 1$):} Unit-normalized relational graph embeddings $\hat{\mathbf{z}}_i \in \mathbb{S}^{d_{\text{graph}}-1}$ learned on $G$ and clustered via KMeans without semantic concatenation.
\end{enumerate}

\subsection{Internal and External Validation Metrics}
\label{subsec:metrics}

We evaluate cluster quality using three internal geometric metrics (requiring no ground-truth labels) and two external ground-truth alignment metrics:
\begin{enumerate}
  \item \textbf{Internal Geometric Validation Metrics:} Let $d(\mathbf{h}_i, \mathbf{h}_j) = 1 - \cos(\mathbf{h}_i, \mathbf{h}_j)$ denote cosine distance for Silhouette evaluation and $\|\mathbf{h}_i - \mathbf{h}_j\|_2$ denote Euclidean distance for centroid dispersion. For each document $d_i \in \mathcal{C}_k$, let $a(i) = \frac{1}{|\mathcal{C}_k|-1}\sum_{d_j \in \mathcal{C}_k \setminus \{d_i\}} d(\mathbf{h}_i, \mathbf{h}_j)$ be the mean intra-cluster distance and $b(i) = \min_{\ell \neq k} \frac{1}{|\mathcal{C}_\ell|}\sum_{d_j \in \mathcal{C}_\ell} d(\mathbf{h}_i, \mathbf{h}_j)$ be the minimum mean inter-cluster distance. The Silhouette Score ($\text{Sil} \in [-1, 1]$, higher is better) \citep{rousseeuw1987silhouettes}, Davies-Bouldin Index ($\text{DBI} \ge 0$, lower is better) \citep{davies1979cluster}, and Calinski-Harabasz Score ($\text{CHS} \ge 0$, higher is better) \citep{calinski1974dendrite} are defined as:
  \begin{equation}
    \text{Sil} = \frac{1}{N} \sum_{i=1}^{N} \frac{b(i) - a(i)}{\max\{a(i), b(i)\}}, \quad
    \text{DBI} = \frac{1}{K} \sum_{k=1}^{K} \max_{\ell \neq k} \frac{\delta_k + \delta_\ell}{\|\mathbf{c}_k - \mathbf{c}_\ell\|_2}, \quad
    \text{CHS} = \frac{\text{Tr}(\mathbf{B}_K) / (K - 1)}{\text{Tr}(\mathbf{W}_K) / (N - K)},
    \label{eq:internal_metrics}
  \end{equation}
  where $\delta_k = \frac{1}{|\mathcal{C}_k|}\sum_{d_i \in \mathcal{C}_k} \|\mathbf{h}_i - \mathbf{c}_k\|_2$ is the intra-cluster scatter of cluster $k$, $\mathbf{B}_K = \sum_{k=1}^K |\mathcal{C}_k|\allowbreak (\mathbf{c}_k - \bar{\mathbf{h}})(\mathbf{c}_k - \bar{\mathbf{h}})^\top$ is the between-cluster dispersion matrix, and $\mathbf{W}_K = \sum_{k=1}^K \sum_{d_i \in \mathcal{C}_k} (\mathbf{h}_i - \mathbf{c}_k)(\mathbf{h}_i - \mathbf{c}_k)^\top$ is the within-cluster dispersion matrix.

  \item \textbf{External Ground-Truth Alignment Metrics:} Given ground-truth legal categories $\mathcal{Y} = \{Y_1, \dots, Y_R\}$ and predicted KMeans clusters $\mathcal{C} = \{\mathcal{C}_1, \dots, \mathcal{C}_K\}$ with contingency counts $n_{rk} = |Y_r \cap \mathcal{C}_k|$, row sums $a_r = |Y_r|$, and column sums $b_k = |\mathcal{C}_k|$, Normalized Mutual Information ($\text{NMI} \in [0, 1]$) \citep{mcdaid2011nmi} and the Adjusted Rand Index ($\text{ARI} \in [-1, 1]$) \citep{hubert1985comparing} quantify label recovery and chance-corrected pairwise agreement:
  \begin{equation}
    \text{NMI}(\mathcal{Y}, \mathcal{C}) = \frac{2 \, I(\mathcal{Y}; \mathcal{C})}{H(\mathcal{Y}) + H(\mathcal{C})} = \frac{2 \sum_{r=1}^R \sum_{k=1}^K \frac{n_{rk}}{N} \log\!\left(\frac{N \, n_{rk}}{a_r b_k}\right)}{-\sum_{r=1}^R \frac{a_r}{N}\log\frac{a_r}{N} - \sum_{k=1}^K \frac{b_k}{N}\log\frac{b_k}{N}},
    \label{eq:nmi_metric}
  \end{equation}
  \begin{equation}
    \text{ARI}(\mathcal{Y}, \mathcal{C}) = \frac{\sum_{r,k} \binom{n_{rk}}{2} - \left[\sum_r \binom{a_r}{2} \sum_k \binom{b_k}{2}\right] / \binom{N}{2}}{\frac{1}{2}\left[\sum_r \binom{a_r}{2} + \sum_k \binom{b_k}{2}\right] - \left[\sum_r \binom{a_r}{2} \sum_k \binom{b_k}{2}\right] / \binom{N}{2}}.
    \label{eq:ari_metric}
  \end{equation}
\end{enumerate}

\subsection{Hyperparameter Search Spaces}
\label{subsec:search_spaces}
We evaluate hyperparameter sensitivity across two complementary regimes:
\begin{enumerate}
  \item \textbf{Full-Factorial Grid Search ($216$ Configurations/Dataset):} Fixing the semantic encoder to all-MiniLM-L6-v2 ($d_{\text{sem}}=384$), $p = q = 1.0$, $w = 5$, and balanced spherical weight $\alpha = 0.5$, we evaluate the full Cartesian grid over $k_{\text{NN}} \in \{2, 5, 10, 15\}$, multi-topic bridges $M \in \{1, 2, 3\}$, graph embedding dimension $d_{\text{graph}} \in \{32, 64, 128\}$, walks per node $R \in \{10, 20, 40\}$, and walk length $L \in \{15, 30\}$ ($4 \times 3 \times 3 \times 3 \times 2 = 216$ runs per dataset; $648$ grid evaluations total).
  \item \textbf{Bayesian Multivariate TPE and Multi-Objective NSGA-II Optimization ($240$ Trials/Dataset):} Let $\boldsymbol{\theta} = (\text{encoder}, T, k_{\text{NN}}, M, \tau, \gamma, \beta, d_{\text{graph}}, L, R, p, q, w, \alpha, \text{norm}, \text{cluster}) \in \Theta$ denote the full 16-dimensional categorical and continuous hyperparameter vector. Using Optuna \citep{akiba2019optuna}, we formulate hyperparameter tuning both as a single-objective Bayesian optimization problem solved via a Multivariate Group-Decomposed Tree-structured Parzen Estimator (TPE, $180$ trials/dataset) \citep{bergstra2011algorithms, falkner2018bohb} and as a multi-objective Pareto frontier search solved via NSGA-II ($60$ trials/dataset) \citep{deb2002nsgaii}:
  \begin{equation}
    \boldsymbol{\theta}^* = \arg\max_{\boldsymbol{\theta} \in \Theta} \, \text{NMI}\big(\mathcal{Y},\, \mathcal{C}(\boldsymbol{\theta})\big), \qquad
    \mathcal{P}^* = \operatorname{ParetoMax}_{\boldsymbol{\theta} \in \Theta} \, \Big( \text{NMI}\big(\mathcal{Y},\, \mathcal{C}(\boldsymbol{\theta})\big), \; \text{Sil}\big(\mathbf{H}(\boldsymbol{\theta}),\, \mathcal{C}(\boldsymbol{\theta})\big) \Big).
    \label{eq:hpo_pareto}
  \end{equation}
\end{enumerate}

\section{Empirical Results and Analysis}
\label{sec:results}

\subsection{Qualitative Manifold Evaluation Across Legal Benchmarks}
\label{subsec:qualitative}

\begin{figure}[!htbp]
  \centering
  \begin{subfigure}[b]{0.46\linewidth}
    \includegraphics[width=\linewidth]{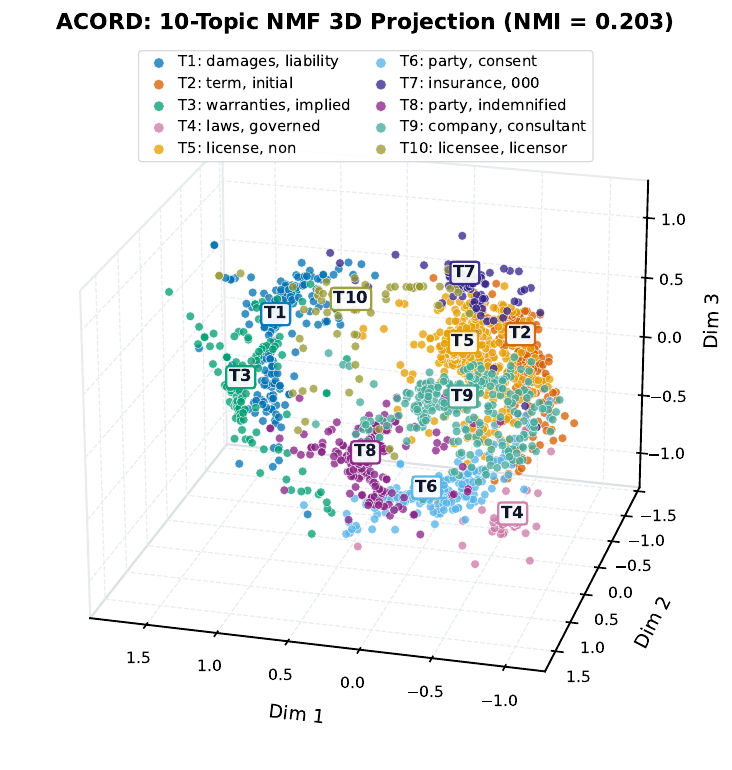}
    \caption{Ten-topic NMF 3D UMAP ($K=10$).}
    \label{fig:nmf_umap_3d}
  \end{subfigure}
  \hfill
  \begin{subfigure}[b]{0.46\linewidth}
    \includegraphics[width=\linewidth]{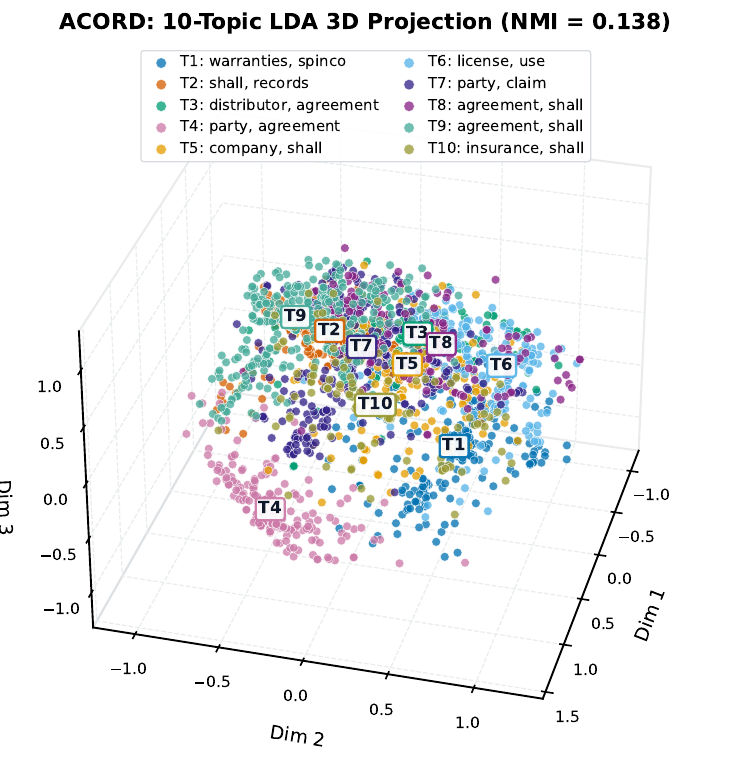}
    \caption{Ten-topic LDA 3D UMAP ($K=10$).}
    \label{fig:lda_umap_3d}
  \end{subfigure}
  \caption{3D UMAP manifold projections for baseline (a) NMF and (b) LDA topic models on the ACORD corpus ($N=2,227, K=10$), annotated with cluster centroid markers (T1--T10) and top representative lexical terms.}
  \label{fig:nmf_lda_umap}
\end{figure}

To contrast classical bag-of-words topic geometry with TGSF's spherical manifold, Figure~\ref{fig:nmf_lda_umap} displays 3D UMAP projections for ten-topic NMF and LDA baselines on ACORD ($N=2,227, K=10$). In the NMF projection (Figure~\ref{fig:nmf_umap_3d}), the ten topic clusters (T1--T10) form localized concentrations for specific lexical clauses (such as governing-law, warranty, and indemnification provisions), whereas in the LDA projection (Figure~\ref{fig:lda_umap_3d}) the simplex topic assignments scatter along diffuse, intersecting trajectories, reflecting the limitations of bag-of-words topic distributions when capturing fine-grained legal phrasing.

\begin{figure}[!htbp]
  \centering
  \begin{subfigure}[b]{0.40\linewidth}
    \includegraphics[width=\linewidth]{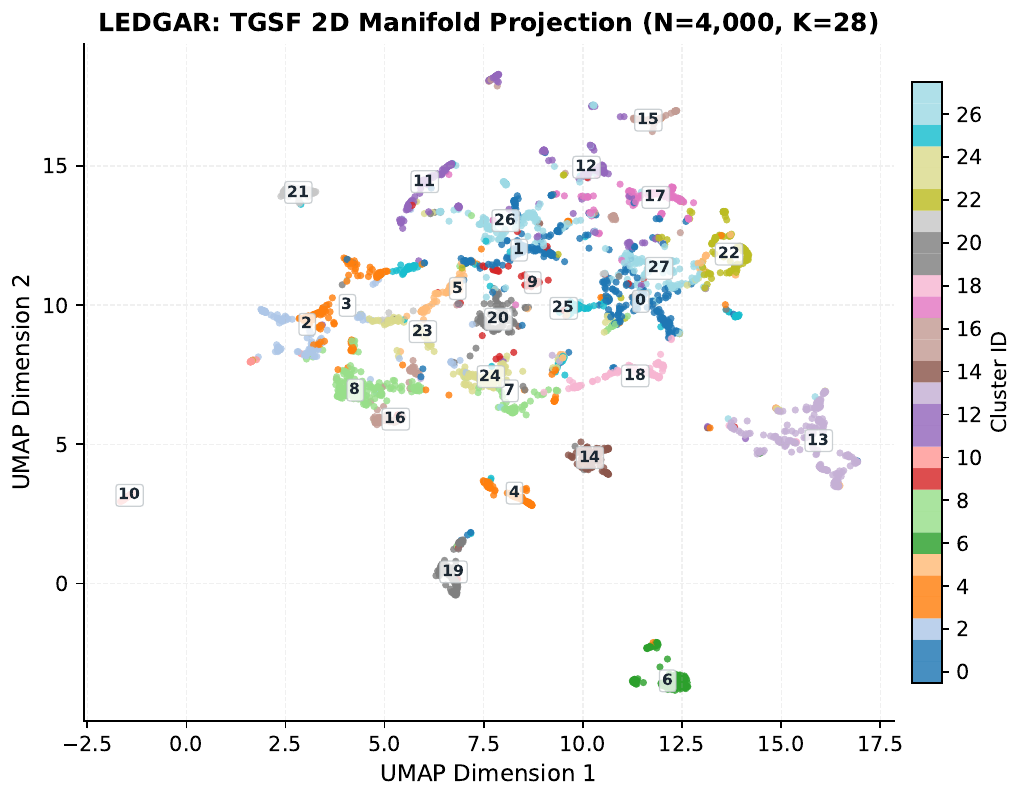}
    \caption{LEDGAR 2D ($N=4,000, K=28$).}
    \label{fig:ledgar_umap_2d}
  \end{subfigure}
  \hspace{0.03\linewidth}
  \begin{subfigure}[b]{0.40\linewidth}
    \includegraphics[width=\linewidth]{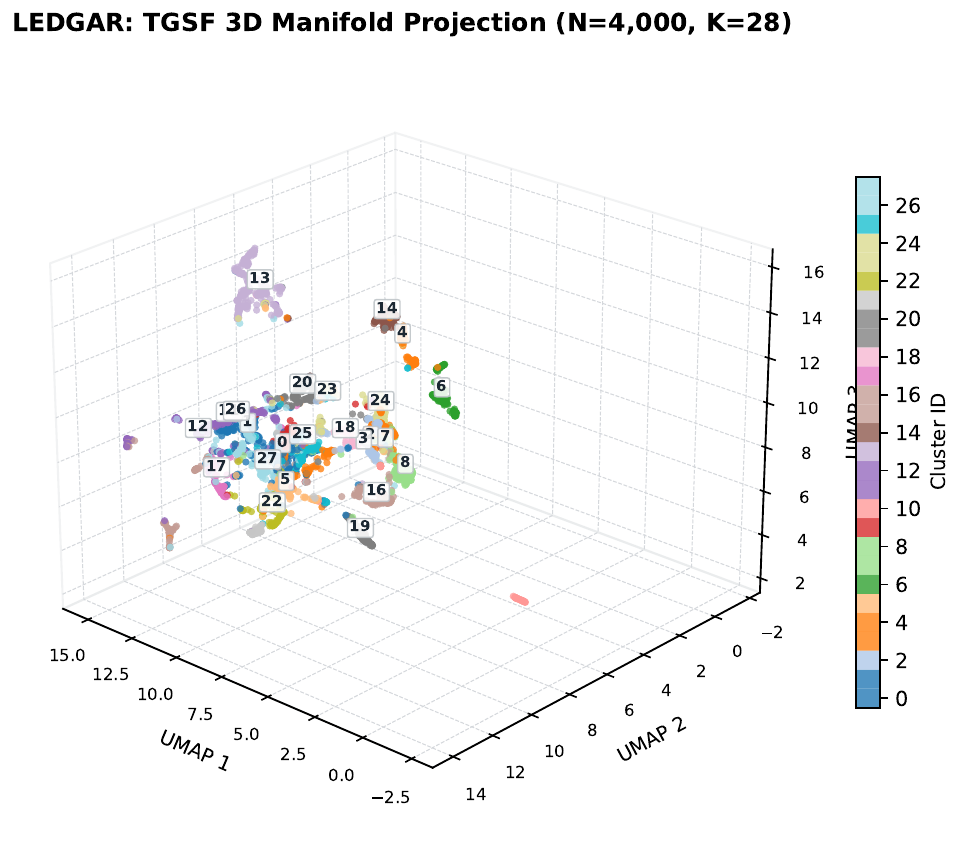}
    \caption{LEDGAR 3D ($N=4,000, K=28$).}
    \label{fig:ledgar_umap_3d}
  \end{subfigure}
  \vspace{1pt}
  \begin{subfigure}[b]{0.40\linewidth}
    \includegraphics[width=\linewidth]{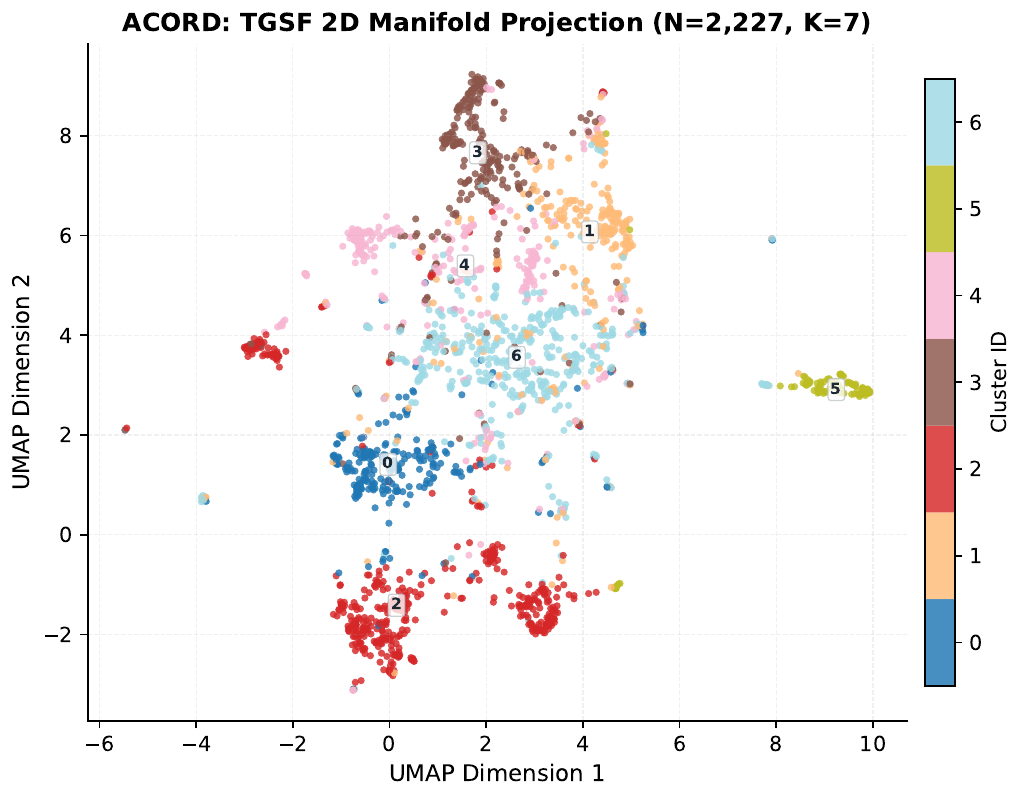}
    \caption{ACORD 2D ($N=2,227, K=7$).}
    \label{fig:acord_umap_2d}
  \end{subfigure}
  \hspace{0.03\linewidth}
  \begin{subfigure}[b]{0.40\linewidth}
    \includegraphics[width=\linewidth]{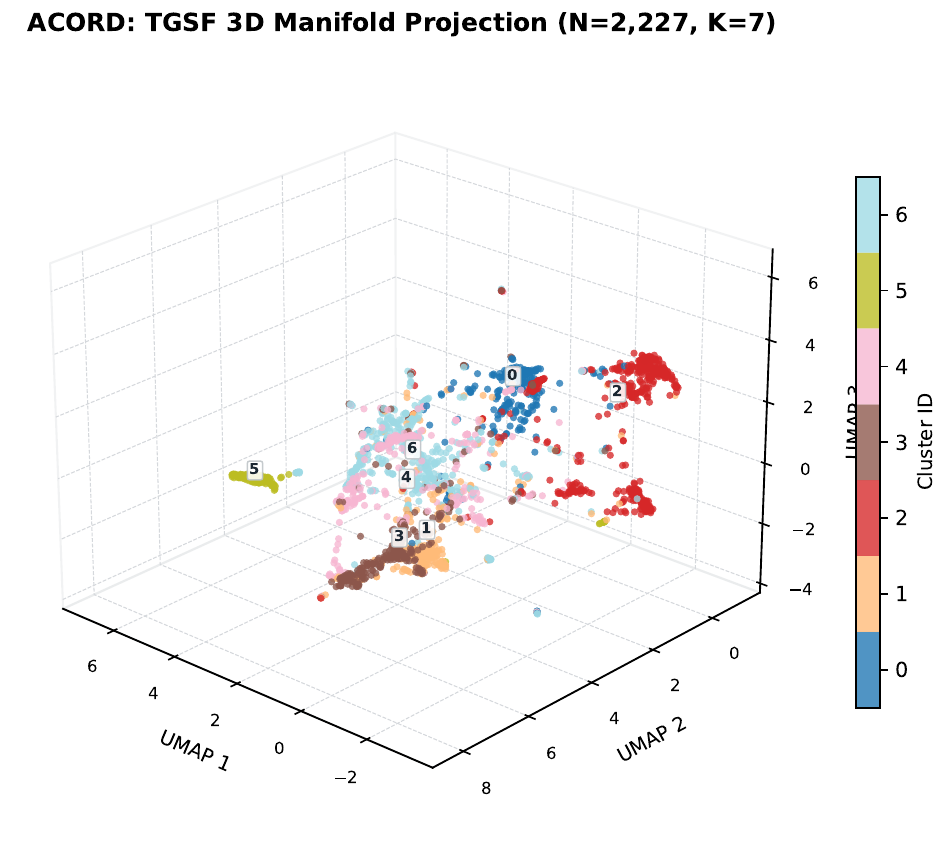}
    \caption{ACORD 3D ($N=2,227, K=7$).}
    \label{fig:acord_umap_3d}
  \end{subfigure}
  \vspace{1pt}
  \begin{subfigure}[b]{0.40\linewidth}
    \includegraphics[width=\linewidth]{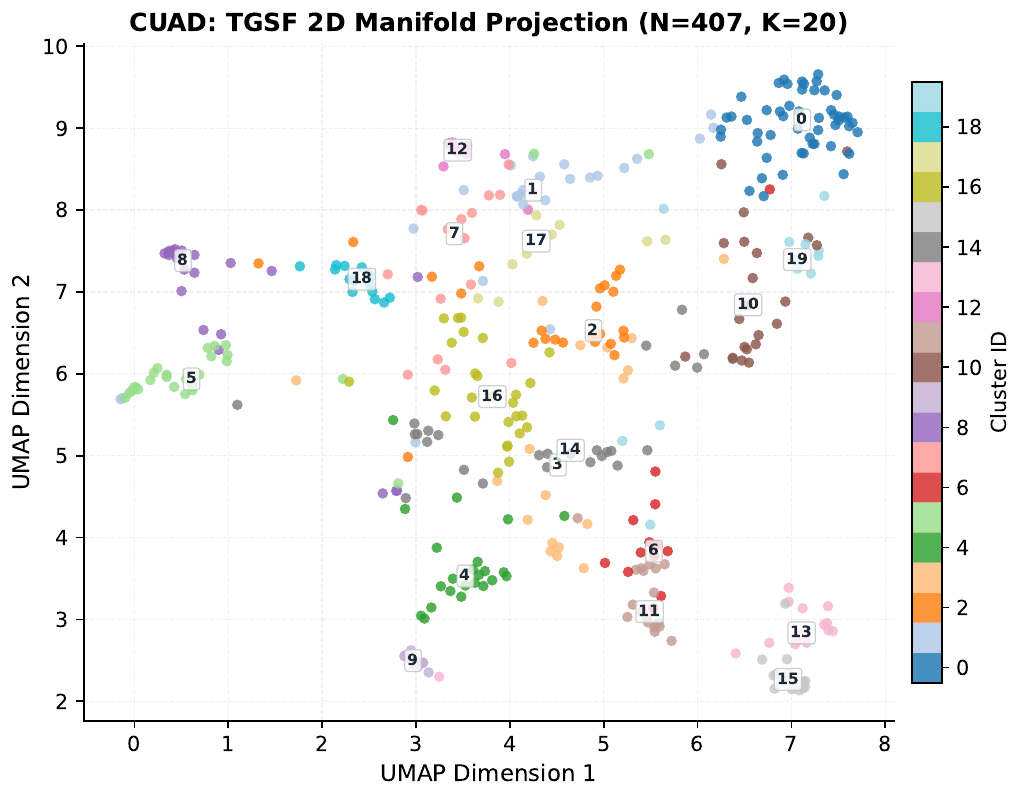}
    \caption{CUAD 2D ($N=407, K=20$).}
    \label{fig:cuad_umap_2d}
  \end{subfigure}
  \hspace{0.03\linewidth}
  \begin{subfigure}[b]{0.40\linewidth}
    \includegraphics[width=\linewidth]{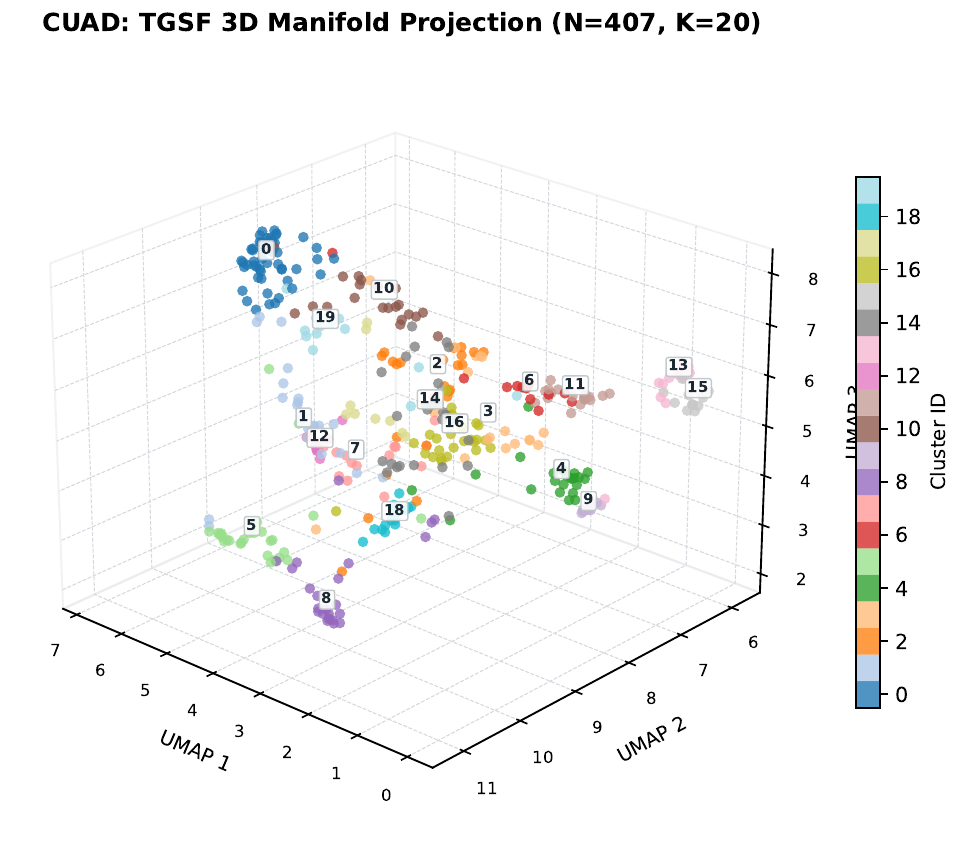}
    \caption{CUAD 3D ($N=407, K=20$).}
    \label{fig:cuad_umap_3d}
  \end{subfigure}
  \caption{Two- and three-dimensional UMAP manifold projections of the TGSF embeddings across LEDGAR ($N=4,000, K=28$), ACORD ($N=2,227, K=7$), and CUAD ($N=407, K=20$). Centroid labels in the 2D projections mark the median coordinates of each cluster.}
  \label{fig:multi_umap}
\end{figure}

\begin{figure}[!htbp]
  \centering
  \includegraphics[width=0.64\textwidth]{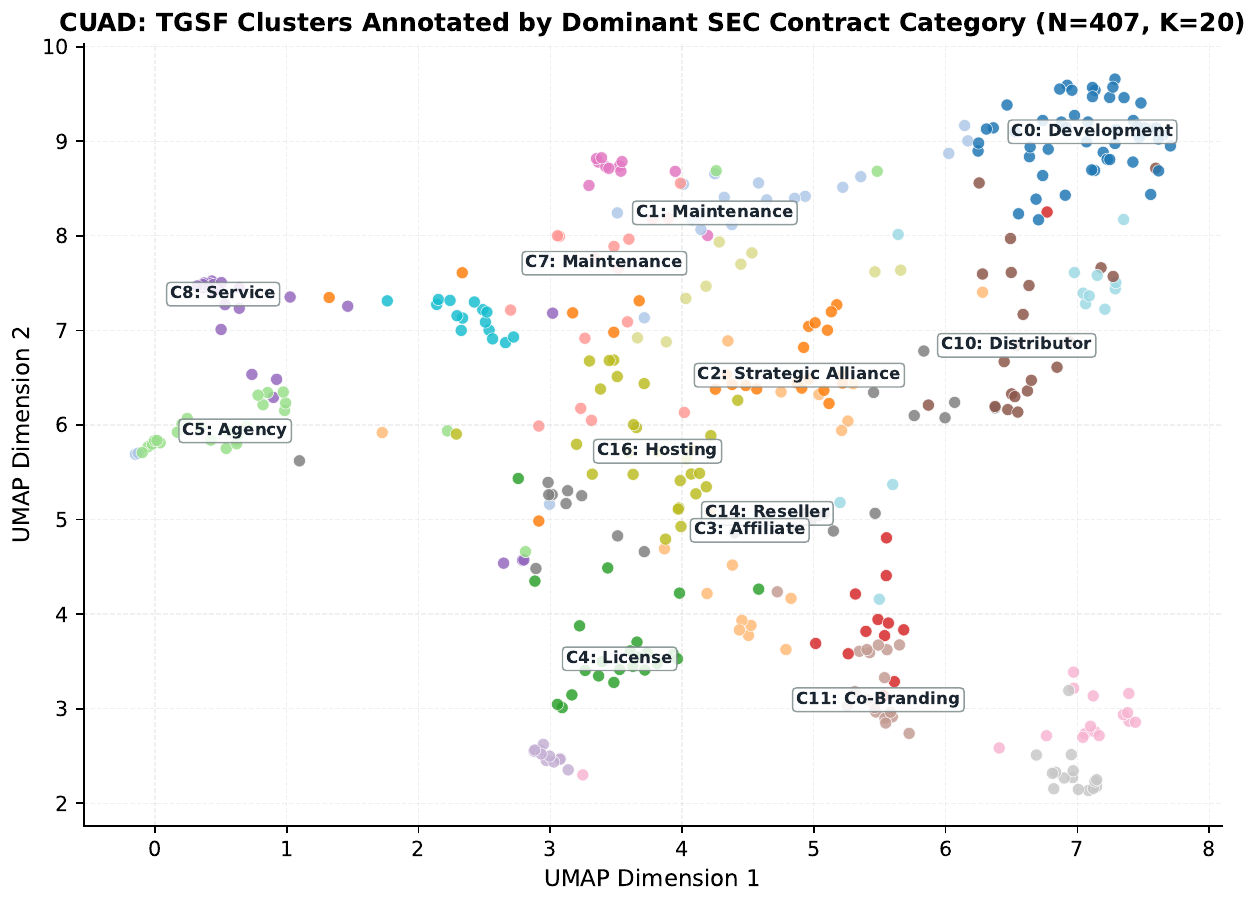}
  \caption{2D UMAP projection of CUAD commercial contract clusters ($N=407, K=20$) generated by TGSF, annotated with the dominant ground-truth SEC contract category for the largest clusters.}
  \label{fig:cuadlabels}
\end{figure}

In contrast, Figure~\ref{fig:multi_umap} presents the 2D and 3D UMAP projections of the learned TGSF embeddings across LEDGAR ($N=4,000, K=28$), ACORD ($N=2,227, K=7$), and CUAD ($N=407, K=20$). Across all three legal benchmarks, the TGSF manifold exhibits well-differentiated, geometrically cohesive cluster regions while preserving smooth local transitions between semantically proximate clauses and contracts. Furthermore, Figure~\ref{fig:cuadlabels} displays the CUAD TGSF manifold annotated with the dominant ground-truth SEC EDGAR contract category for each major cluster (including Outsourcing, Service, License, Distributor, Development, Marketing, and Strategic Alliance), demonstrating that the unsupervised clusters align with practitioner-recognized agreement types during contract review.

\subsection{Multi-Dataset Quantitative Benchmark Comparison}
\label{subsec:quantitative}

\begin{figure}[!htbp]
  \centering
  \includegraphics[width=0.90\textwidth]{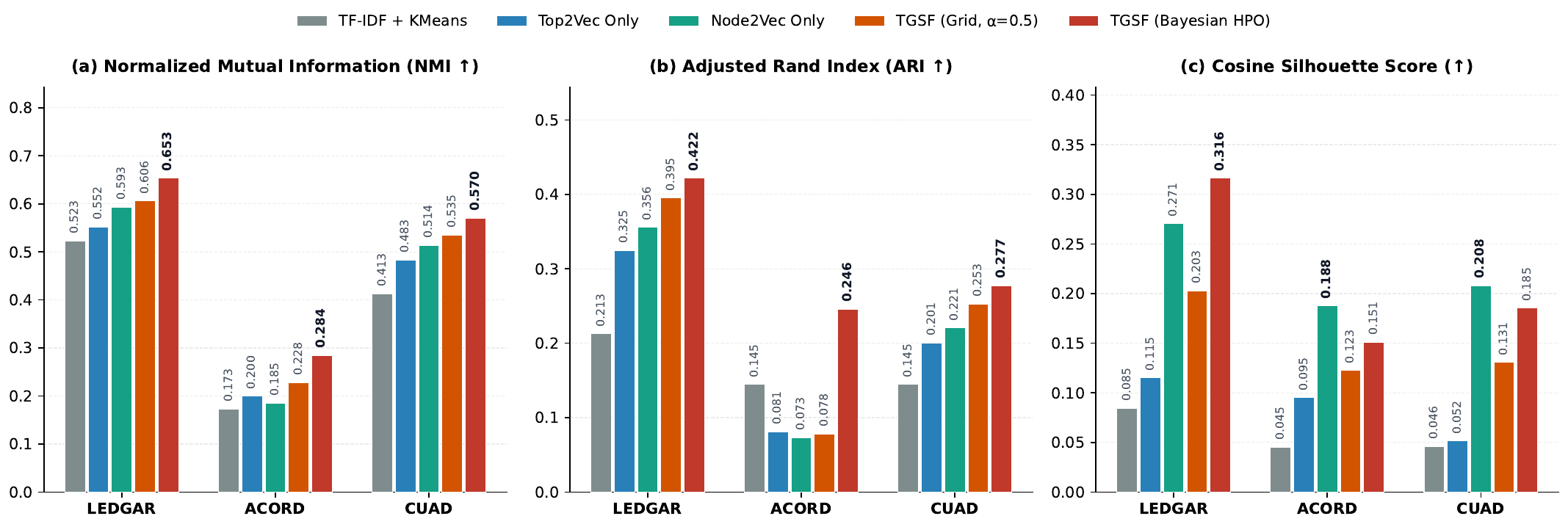}
  \caption{Cross-dataset comparison of (a) Normalized Mutual Information (NMI), (b) Adjusted Rand Index (ARI), and (c) Cosine Silhouette Score across TF-IDF + KMeans, Top2Vec Only (Semantic, $\alpha=0$), Node2Vec Only (Structural, $\alpha=1$), TGSF (Grid, $\alpha=0.5$), and TGSF (Bayesian HPO) on LEDGAR, ACORD, and CUAD.}
  \label{fig:cross_dataset}
\end{figure}

Table~\ref{tab:multi_dataset_benchmark} and Figure~\ref{fig:cross_dataset} report the comprehensive quantitative benchmark comparison across LEDGAR, ACORD, and CUAD. Three principal findings emerge:
\begin{enumerate}
  \item \textbf{Superior Ground-Truth Category Recovery on LEDGAR:} On the LEDGAR provision benchmark ($N=4,000, K=28$), TGSF achieves the highest external validation scores under the all-MiniLM-L6-v2 grid ($\text{NMI} = 0.606, \text{ARI} = 0.395$), and under Bayesian TPE optimization with Legal-BERT reaches $\text{NMI} = 0.653$ and $\text{ARI} = 0.422$ (up to $\text{ARI} = 0.425$). This corresponds to a $+24.9\%$ relative improvement in NMI and a $+98.1\%$ relative improvement in ARI over TF-IDF + KMeans ($\text{NMI} = 0.523, \text{ARI} = 0.213$), outperforming LDA ($\text{NMI} = 0.567, \text{ARI} = 0.331$), NMF ($\text{NMI} = 0.509, \text{ARI} = 0.176$), Top2Vec-only ($\text{NMI} = 0.553, \text{ARI} = 0.324$), and Node2Vec-only ($\text{NMI} = 0.593, \text{ARI} = 0.356$).
  \item \textbf{Consistent Gains Across Clause and Full-Contract Granularities (ACORD and CUAD):} On ACORD ($N=2,227, K=7$), TGSF (Grid) achieves $\text{NMI} = 0.228$, and TGSF (Bayesian HPO with Legal-BERT) elevates external alignment to $\text{NMI} = 0.284$ and $\text{ARI} = 0.246$ (up to $\text{ARI} = 0.260$), exceeding all lexical, probabilistic, matrix-factorization, and unimodal baselines on both NMI and ARI (including NMF's $\text{NMI} = 0.183, \text{ARI} = 0.175$). Similarly, on full commercial contracts in CUAD ($N=407, K=20$), TGSF improves from $\text{NMI} = 0.535, \text{ARI} = 0.253$ (Grid) to $\text{NMI} = 0.570, \text{ARI} = 0.277$ (up to $\text{ARI} = 0.282$) under Bayesian HPO with all-mpnet-base-v2, outperforming TF-IDF ($\text{NMI} = 0.413, \text{ARI} = 0.145$) by $+38.0\%$ in NMI and $+91.0\%$ in ARI.
  \item \textbf{Simplex Corner Concentration vs.\ High-Dimensional Spherical Manifolds:} As shown in Table~\ref{tab:multi_dataset_benchmark}, LDA and NMF exhibit numerically high internal Silhouette and CHS values because each document is represented as a low-dimensional $K$-simplex vector ($K \in \{7, 20, 28\}$) concentrated near simplex vertices, whereas TGSF embeds documents on a continuous $(d_{\text{sem}} + d_{\text{graph}})$-dimensional unit hypersphere ($416\text{--}1,024$ dimensions). Crucially, simplex corner concentration does not translate into accurate legal category recovery. Moreover, jointly optimizing the encoder backbone, second-order walk biases $(p, q)$, edge sharpening $(\beta, \gamma)$, and spherical fusion weight $\alpha$ via Bayesian TPE simultaneously improves internal Silhouette over the fixed $\alpha=0.5$ grid ($0.203 \to 0.316$ on LEDGAR, $0.123 \to 0.151$ on ACORD, and $0.131 \to 0.185$ on CUAD) while maximizing external legal category alignment.
\end{enumerate}

\begin{table}[!htbp]
  \centering
  \caption{Multi-Dataset Benchmark Comparison Across LEDGAR ($N=4,000, K=28$), ACORD ($N=2,227, K=7$), and CUAD ($N=407, K=20$) on the Connected Graph $G = (V_D \cup V_T, E_{DT} \cup E_{DD}, W)$. Best external validation scores per dataset are highlighted in bold.}
  \label{tab:multi_dataset_benchmark}
  \resizebox{\linewidth}{!}{%
  \begin{tabular}{@{}llrrrrr@{}}
    \toprule
    \textbf{Dataset} & \textbf{Model Configuration} & \textbf{Silhouette ($\uparrow$)} & \textbf{DBI ($\downarrow$)} & \textbf{CHS ($\uparrow$)} & \textbf{NMI ($\uparrow$)} & \textbf{ARI ($\uparrow$)} \\
    \midrule
    LEDGAR ($N=4,000, K=28$)
      & TF-IDF + KMeans                  & 0.085 & 4.564 &   25.9 & 0.523 & 0.213 \\
      & LDA ($K=28$)                     & 0.594 & 1.138 &  601.0 & 0.567 & 0.331 \\
      & NMF ($K=28$)                     & 0.336 & 1.041 &  439.4 & 0.509 & 0.176 \\
      & Top2Vec Only (Semantic, $\alpha=0$) & 0.117 & 3.063 &   61.2 & 0.553 & 0.324 \\
      & Node2Vec Only (Structural $G$, $\alpha=1$) & 0.271 & 2.015 &  110.5 & 0.593 & 0.356 \\
      & TGSF (Grid Search, MiniLM, $\alpha=0.5$) & 0.203 & 2.421 &   88.1 & 0.606 & 0.395 \\
      & TGSF (Bayesian HPO, Legal-BERT, $\alpha=0.83$) & 0.316 & 1.916 &  137.2 & \textbf{0.653} & \textbf{0.422}$^\dagger$ \\
    \midrule
    ACORD ($N=2,227, K=7$)
      & TF-IDF + KMeans                  & 0.045 & 5.084 &   26.6 & 0.173 & 0.145 \\
      & LDA ($K=7$)                      & 0.713 & 0.700 & 1569.4 & 0.143 & 0.036 \\
      & NMF ($K=7$)                      & 0.360 & 0.679 & 1108.2 & 0.183 & 0.175 \\
      & Top2Vec Only (Semantic, $\alpha=0$) & 0.095 & 3.480 &   76.5 & 0.200 & 0.081 \\
      & Node2Vec Only (Structural $G$, $\alpha=1$) & 0.188 & 2.668 &  119.3 & 0.185 & 0.073 \\
      & TGSF (Grid Search, MiniLM, $\alpha=0.5$) & 0.123 & 3.263 &   89.7 & 0.228 & 0.115$^*$ \\
      & TGSF (Bayesian HPO, Legal-BERT, $\alpha=0.19$) & 0.151 & 2.384 &  118.9 & \textbf{0.284} & \textbf{0.246}$^\dagger$ \\
    \midrule
    CUAD ($N=407, K=20$)
      & TF-IDF + KMeans                  & 0.046 & 4.757 &    2.7 & 0.413 & 0.145 \\
      & LDA ($K=20$)                     & 0.746 & 0.621 &  110.7 & 0.320 & 0.063 \\
      & NMF ($K=20$)                     & 0.669 & 1.034 &   45.0 & 0.506 & 0.232 \\
      & Top2Vec Only (Semantic, $\alpha=0$) & 0.052 & 3.469 &    5.2 & 0.483 & 0.201 \\
      & Node2Vec Only (Structural $G$, $\alpha=1$) & 0.208 & 2.098 &   15.5 & 0.514 & 0.221 \\
      & TGSF (Grid Search, MiniLM, $\alpha=0.5$) & 0.131 & 2.667 &    9.8 & 0.535 & 0.253 \\
      & TGSF (Bayesian HPO, MPNet, $\alpha=0.77$) & 0.185 & 2.306 &   15.6 & \textbf{0.570} & \textbf{0.277}$^\dagger$ \\
    \bottomrule
  \end{tabular}%
  }
  \vspace{2pt}
  \footnotesize{\\ $^*$On ACORD ($K=7$) under the all-MiniLM-L6-v2 grid, the NMI-optimal configuration ($k_{\text{NN}}=15, M=3, d_{\text{graph}}=64$) achieves $\text{NMI}=0.228, \text{ARI}=0.078$, whereas the ARI-optimal configuration ($k_{\text{NN}}=2, M=1, d_{\text{graph}}=32$) achieves $\text{ARI}=0.115, \text{NMI}=0.213$. $^\dagger$Across the 240-trial Bayesian TPE + Pareto studies, the ARI-maximizing trials reach $\text{ARI} = 0.425$ ($\text{NMI}=0.649$) on LEDGAR, $\text{ARI} = 0.260$ ($\text{NMI}=0.284$) on ACORD, and $\text{ARI} = 0.282$ ($\text{NMI}=0.567$) on CUAD.}
\end{table}

\subsection{Factorial Grid Sensitivity, Cluster-Count Robustness, and Corpus-Volume Saturation}
\label{subsec:sensitivity}

To examine the stability of TGSF across operating regimes, we evaluated sensitivity across the 216-configuration factorial hyperparameter grid, target cluster counts $K \in [5, 30]$, and corpus subsampling fractions from $50\%$ to $100\%$.

\begin{figure}[!htbp]
  \centering
  \begin{subfigure}[b]{0.82\linewidth}
    \includegraphics[width=\linewidth]{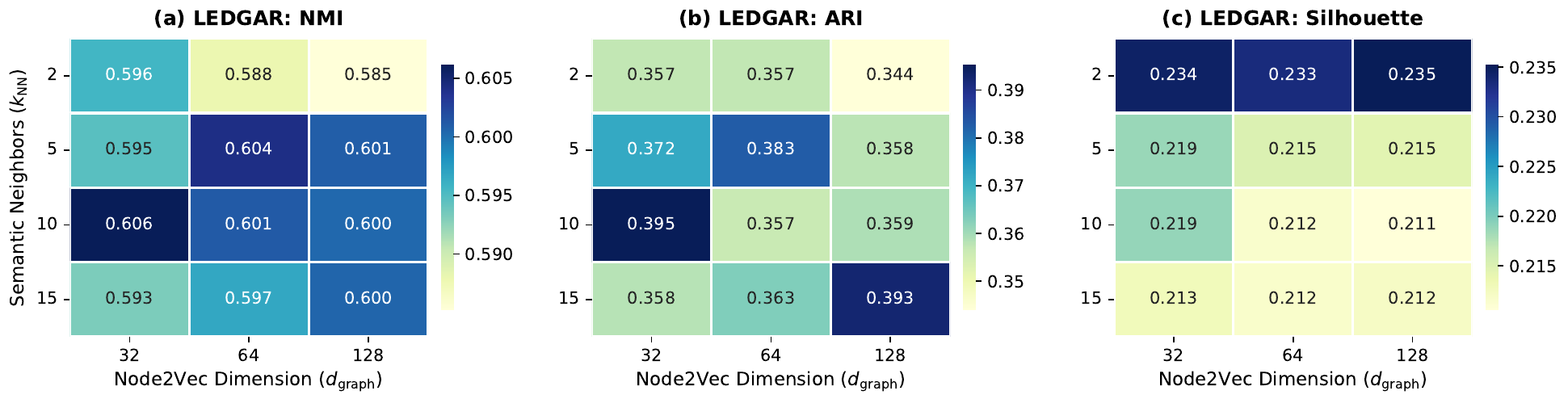}
    \caption{LEDGAR hyperparameter sensitivity ($k_{\text{NN}}$ vs.\ $d_{\text{graph}}$).}
    \label{fig:ledgar_heatmap}
  \end{subfigure}
  \vspace{1pt}
  \begin{subfigure}[b]{0.82\linewidth}
    \includegraphics[width=\linewidth]{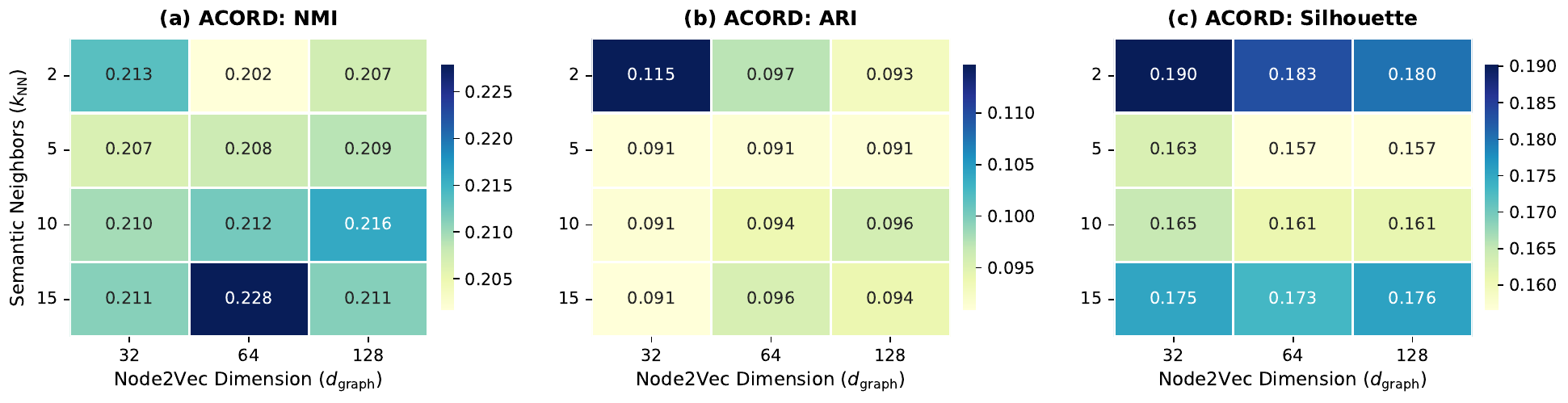}
    \caption{ACORD hyperparameter sensitivity ($k_{\text{NN}}$ vs.\ $d_{\text{graph}}$).}
    \label{fig:acord_heatmap}
  \end{subfigure}
  \vspace{1pt}
  \begin{subfigure}[b]{0.82\linewidth}
    \includegraphics[width=\linewidth]{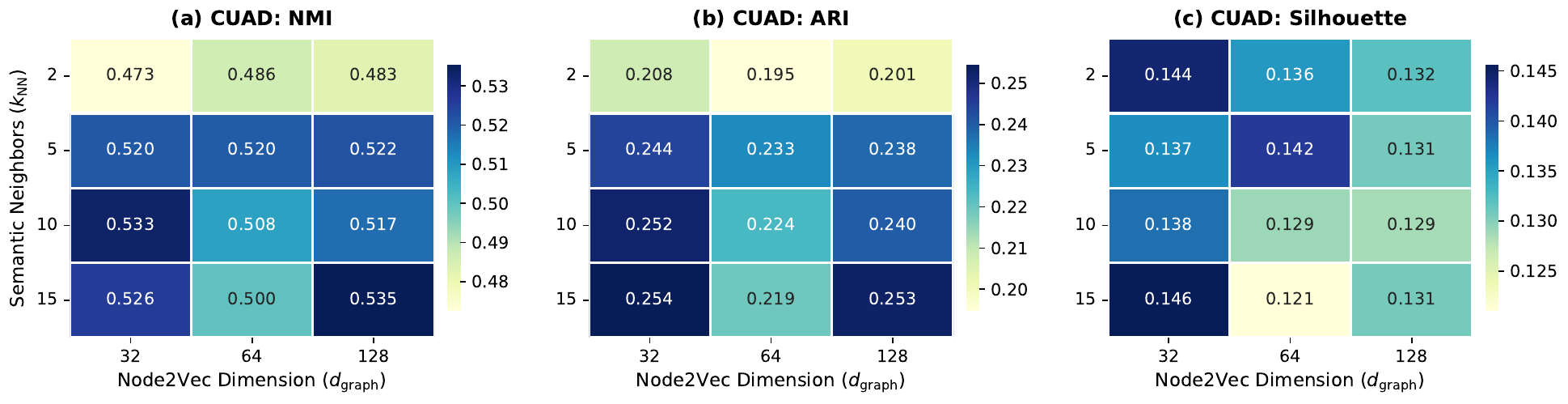}
    \caption{CUAD hyperparameter sensitivity ($k_{\text{NN}}$ vs.\ $d_{\text{graph}}$).}
    \label{fig:cuad_heatmap}
  \end{subfigure}
  \caption{Hyperparameter sensitivity heatmaps illustrating the interaction between semantic neighborhood size ($k_{\text{NN}} \in \{2, 5, 10, 15\}$) and Node2Vec embedding dimensionality ($d_{\text{graph}} \in \{32, 64, 128\}$) across (a) LEDGAR, (b) ACORD, and (c) CUAD.}
  \label{fig:multi_heatmaps}
\end{figure}

\paragraph{Factorial Grid Stability ($216$ Configurations/Dataset).} Table~\ref{tab:grid_summary} reports the distributional stability (mean $\pm$ standard deviation, min--max range, and optimal configuration) across all $216$ grid runs per dataset, and Figure~\ref{fig:multi_heatmaps} visualizes the interaction between $k_{\text{NN}}$ and $d_{\text{graph}}$ across LEDGAR, ACORD, and CUAD. TGSF exhibits high stability across all hyperparameter combinations: on LEDGAR, NMI has a mean of $0.579$ with a standard deviation of only $0.011$ across all $216$ runs, meaning that even the lowest-scoring grid configuration ($\text{NMI} = 0.546$) outperforms the TF-IDF baseline ($\text{NMI} = 0.523$). Across all three datasets, sparse $k$-NN connectivity ($k_{\text{NN}} = 2$) yields higher local Silhouette scores ($0.235$ on LEDGAR, $0.190$ on ACORD, $0.144$ on CUAD) by restricting inter-cluster walk transitions, whereas increasing neighborhood connectivity to $k_{\text{NN}} \in \{10, 15\}$ enables random walks to smoothly traverse semantically proximate clauses and contracts, achieving peak grid NMI across LEDGAR ($0.606$ at $k_{\text{NN}}=10$), ACORD ($0.228$ at $k_{\text{NN}}=15$), and CUAD ($0.535$ at $k_{\text{NN}}=15$).

\begin{table}[!htbp]
  \centering
  \caption{Distributional Stability of TGSF Across the 216-Configuration Hyperparameter Grid Search ($k_{\text{NN}} \in \{2,5,10,15\}$, $M \in \{1,2,3\}$, $d_{\text{graph}} \in \{32,64,128\}$, $R \in \{10,20,40\}$, $L \in \{15,30\}$).}
  \label{tab:grid_summary}
  \resizebox{\linewidth}{!}{%
  \begin{tabular}{@{}llccccc@{}}
    \toprule
    \textbf{Dataset} & \textbf{Statistic} & \textbf{NMI} & \textbf{ARI} & \textbf{Silhouette} & \textbf{DBI} & \textbf{CHS} \\
    \midrule
    LEDGAR ($N=4,000, K=28$)
      & Mean $\pm$ SD & $0.579 \pm 0.011$ & $0.340 \pm 0.016$ & $0.191 \pm 0.019$ & $2.471 \pm 0.125$ & $78.6 \pm 6.7$ \\
      & Min -- Max    & $0.546 \text{ -- } 0.606$ & $0.297 \text{ -- } 0.395$ & $0.156 \text{ -- } 0.235$ & $2.248 \text{ -- } 2.847$ & $66.3 \text{ -- } 96.6$ \\
      & Optimal Config & \multicolumn{5}{l}{$k_{\text{NN}}=10,\; M=1,\; d_{\text{graph}}=32,\; R=40,\; L=15$} \\
    \midrule
    ACORD ($N=2,227, K=7$)
      & Mean $\pm$ SD & $0.198 \pm 0.011$ & $0.078 \pm 0.015$ & $0.151 \pm 0.017$ & $3.189 \pm 0.263$ & $99.8 \pm 11.4$ \\
      & Min -- Max    & $0.165 \text{ -- } 0.228$ & $0.036 \text{ -- } 0.115$ & $0.114 \text{ -- } 0.190$ & $2.627 \text{ -- } 4.005$ & $75.4 \text{ -- } 129.5$ \\
      & Optimal Config & \multicolumn{5}{l}{$k_{\text{NN}}=15,\; M=3,\; d_{\text{graph}}=64,\; R=10,\; L=30$} \\
    \midrule
    CUAD ($N=407, K=20$)
      & Mean $\pm$ SD & $0.511 \pm 0.019$ & $0.230 \pm 0.020$ & $0.135 \pm 0.007$ & $2.768 \pm 0.142$ & $9.1 \pm 0.8$ \\
      & Min -- Max    & $0.473 \text{ -- } 0.535$ & $0.195 \text{ -- } 0.254$ & $0.121 \text{ -- } 0.146$ & $2.580 \text{ -- } 3.012$ & $7.8 \text{ -- } 10.4$ \\
      & Optimal Config & \multicolumn{5}{l}{$k_{\text{NN}}=15,\; M=1,\; d_{\text{graph}}=128,\; R=20,\; L=15$} \\
    \bottomrule
  \end{tabular}%
  }
\end{table}

\begin{figure}[!htbp]
  \centering
  \begin{subfigure}[b]{0.44\linewidth}
    \includegraphics[width=\linewidth]{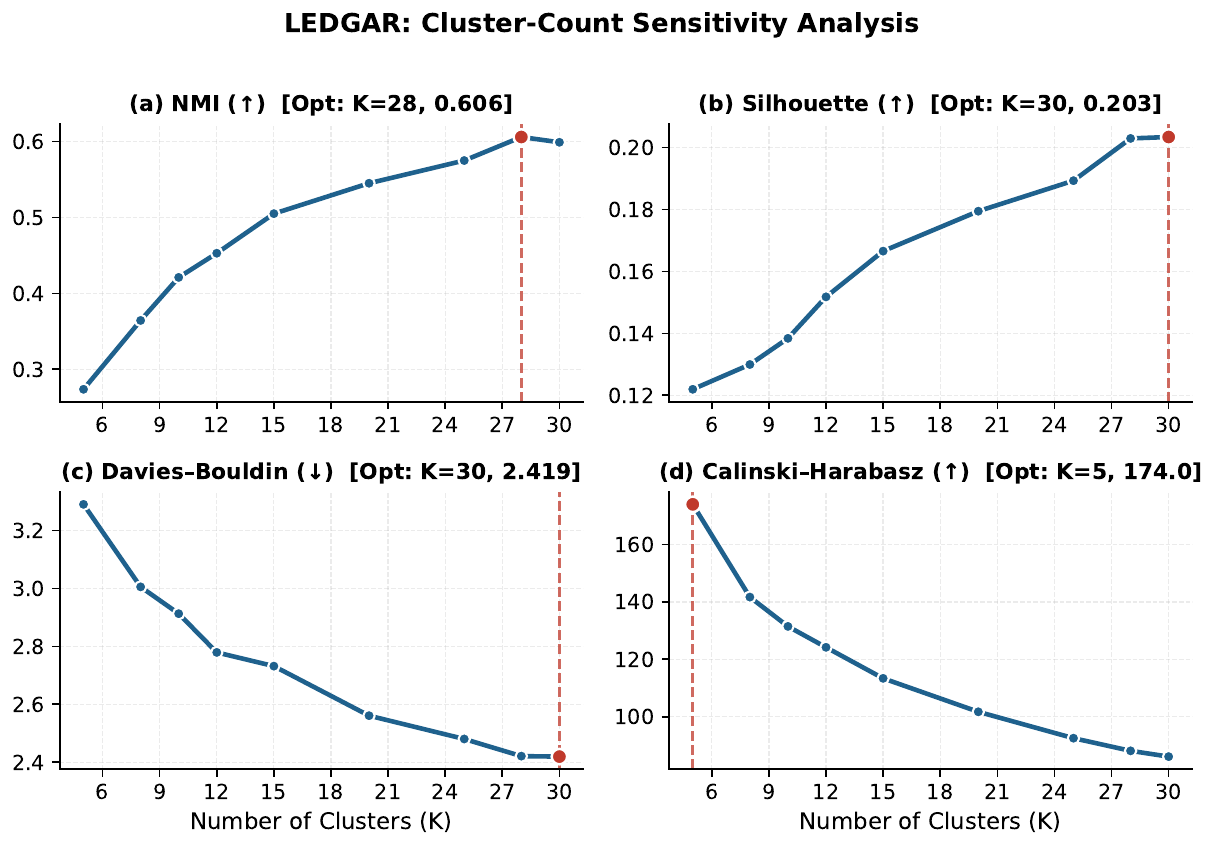}
    \caption{LEDGAR sensitivity ($K \in [5, 30]$).}
    \label{fig:ledgar_sens}
  \end{subfigure}
  \hspace{0.03\linewidth}
  \begin{subfigure}[b]{0.44\linewidth}
    \includegraphics[width=\linewidth]{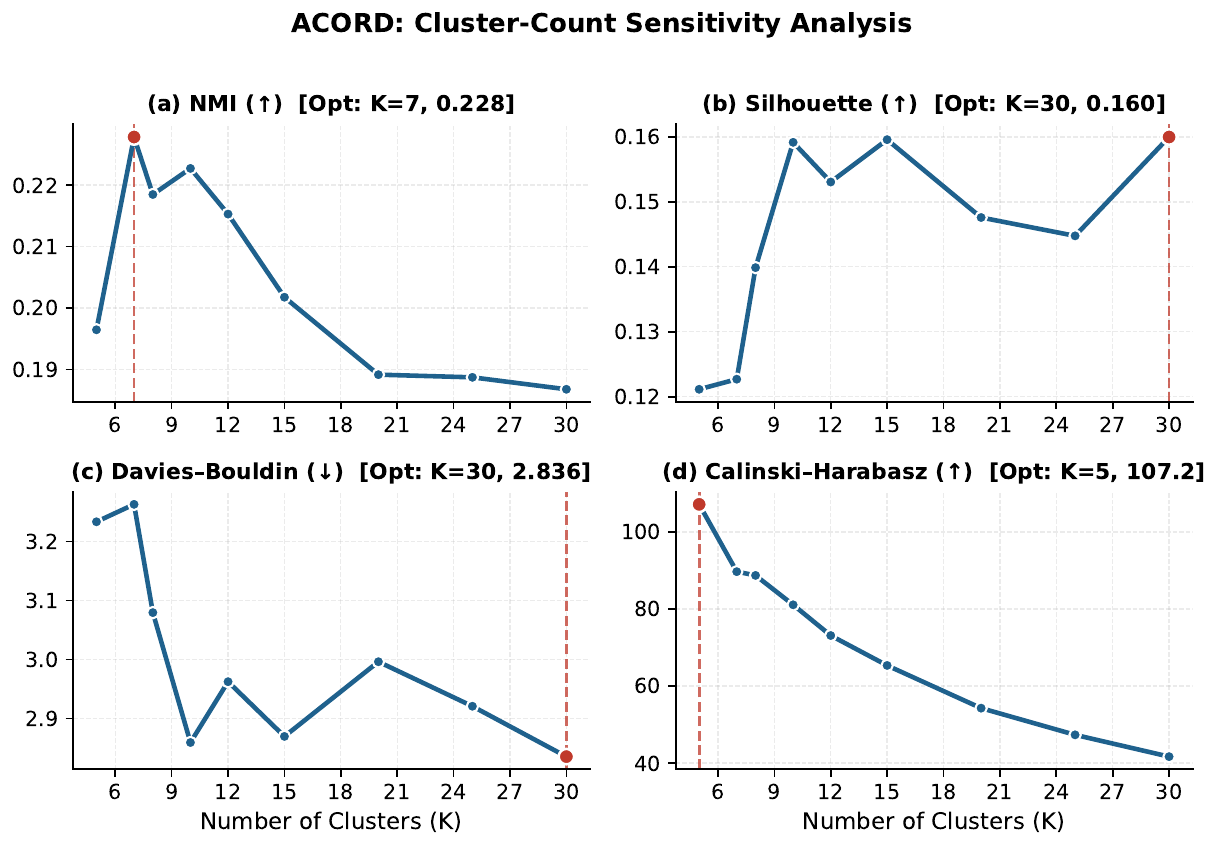}
    \caption{ACORD sensitivity ($K \in [5, 30]$).}
    \label{fig:acord_sens}
  \end{subfigure}
  \vspace{2pt}
  \begin{subfigure}[b]{0.44\linewidth}
    \includegraphics[width=\linewidth]{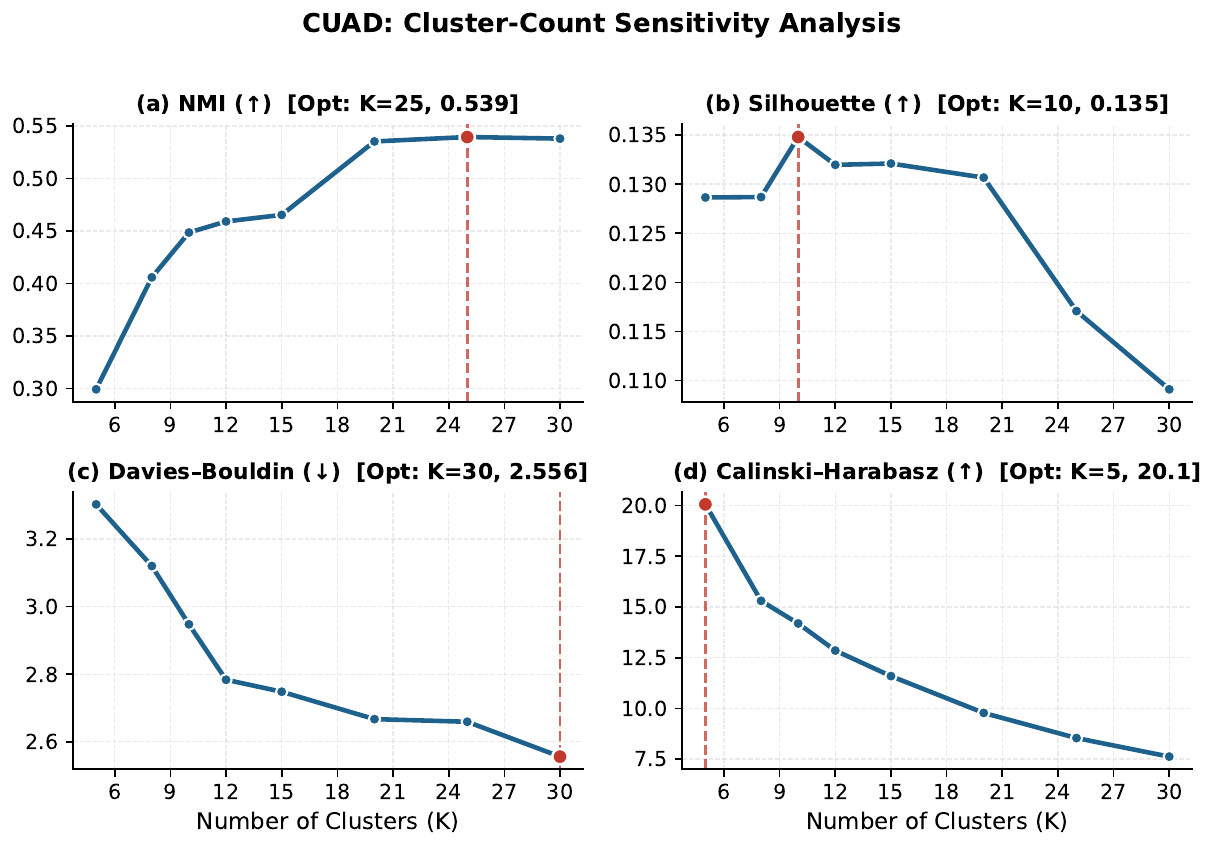}
    \caption{CUAD sensitivity ($K \in [5, 30]$).}
    \label{fig:cuad_sens}
  \end{subfigure}
  \caption{Multi-dataset sensitivity analysis of TGSF across cluster counts $K \in \{5, 8, 10, 12, 15, 20, 25, 28, 30\}$ for (a) LEDGAR, (b) ACORD, and (c) CUAD, with vertical dashed markers indicating the optimal $K^*$ for each metric.}
  \label{fig:multi_sensitivity}
\end{figure}

\begin{figure}[!htbp]
  \centering
  \includegraphics[width=0.76\textwidth]{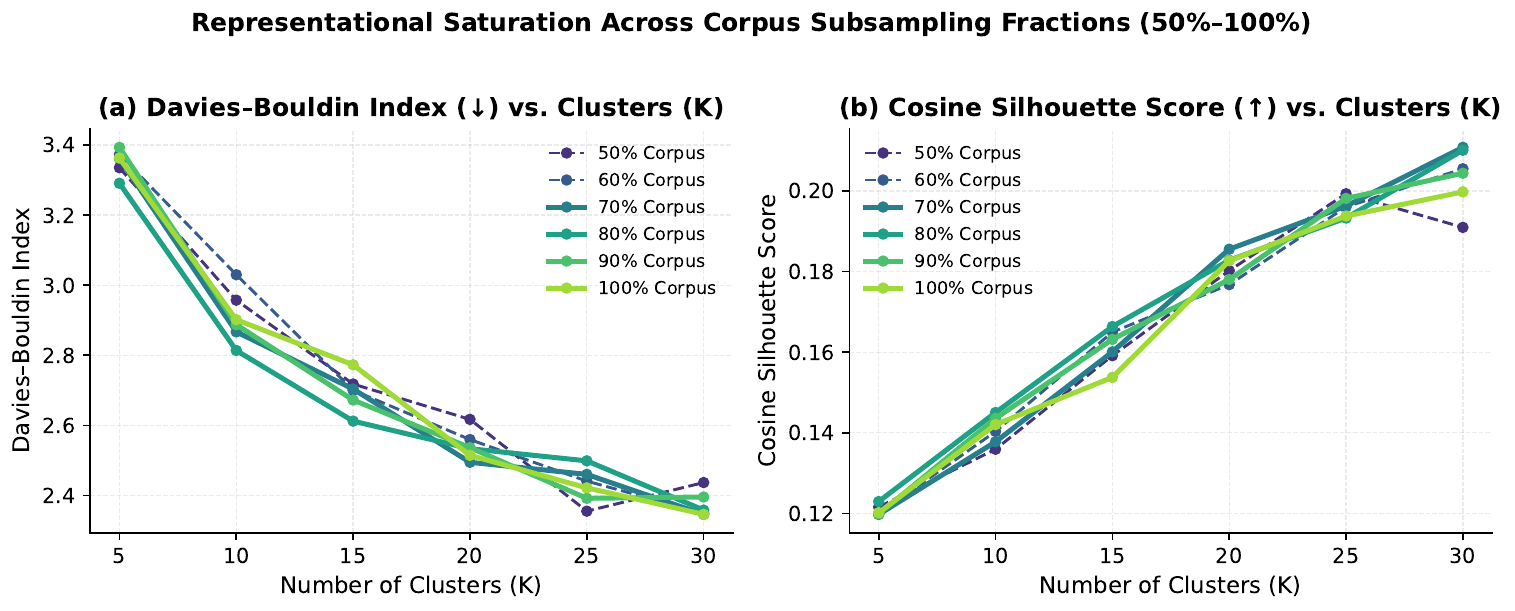}
  \caption{Representational saturation analysis showing (a) Davies--Bouldin Index (lower is better) and (b) Cosine Silhouette Score (higher is better) versus number of clusters $K$ across corpus subsampling fractions ($50\%\text{--}100\%$). Subsets $\ge 70\%$ (solid lines) closely track the full corpus trajectory.}
  \label{fig:DB_S_metrics}
\end{figure}

\paragraph{Sensitivity to Target Cluster Count ($K$) and Corpus Volume.} Figure~\ref{fig:multi_sensitivity} presents the multi-metric sensitivity curves across $K \in \{5, 8, 10, 12, 15, 20, 25, 28, 30\}$ for LEDGAR, ACORD, and CUAD:
\begin{itemize}
  \item On LEDGAR (Figure~\ref{fig:ledgar_sens}), NMI increases monotonically from $0.274$ ($K=5$) to a peak of $0.606$ (and $\text{ARI} = 0.395$) at $K=28$ before plateauing at $K=30$ ($\text{NMI} = 0.599$), matching the true number of functional categories while Silhouette rises from $0.122$ to $0.203$ and DBI drops from $3.290$ to $2.419$.
  \item On ACORD (Figure~\ref{fig:acord_sens}), when evaluated against the $7$ ground-truth macro-categories, NMI peaks in the coarser regime ($K=5\text{--}7$, with $\text{NMI} = 0.228$ at $K=7$) and gradually declines as macro-categories are subdivided into fine-grained sub-clauses, whereas internal geometric compactness (Silhouette and DBI) continues to improve up to $K=15\text{--}30$ (reflecting the $25$ fine-grained ACORD clause queries).
  \item On CUAD (Figure~\ref{fig:cuad_sens}), NMI increases steadily as $K$ approaches the number of underlying SEC contract categories ($24$), reaching $\text{NMI} = 0.535$ ($\text{ARI} = 0.253$) at $K=20$ and peaking at $\text{NMI} = 0.539$ at $K=25$, while DBI decreases monotonically down to $K=30$ ($2.556$).
\end{itemize}
Furthermore, Figure~\ref{fig:DB_S_metrics} displays Davies--Bouldin and Cosine Silhouette curves across corpus subsampling fractions ($50\%, 60\%, 70\%, 80\%, 90\%, \text{ and } 100\%$). Performance curves for subsets of $70\%$ and greater converge closely onto the full $100\%$ corpus trajectory, demonstrating representational saturation: a $70\text{--}80\%$ stratified subsample yields nearly identical cluster geometry while reducing $O(N^2)$ similarity computation and random-walk overhead by $30\text{--}40\%$.

\subsection{Bayesian Multivariate TPE and Multi-Objective NSGA-II Pareto Optimization}
\label{subsec:hpo}

To move beyond discrete grid search and systematically explore the continuous and categorical 16-dimensional hyperparameter space $\boldsymbol{\theta} \in \Theta$ (Equation~\eqref{eq:hpo_pareto}), we executed $240$ GPU-accelerated optimization trials per dataset ($180$ Multivariate Group-Decomposed TPE trials \citep{bergstra2011algorithms, falkner2018bohb} followed by $60$ Multi-Objective NSGA-II Pareto frontier trials \citep{deb2002nsgaii}, totaling $720$ trials across LEDGAR, ACORD, and CUAD). In the multi-objective phase, NSGA-II approximates the non-dominated Pareto set $\mathcal{P}^*$:
\begin{equation}
  \begin{aligned}
    \mathcal{P}^* = \Big\{ \boldsymbol{\theta} \in \Theta \;\Big|\; &\nexists\, \boldsymbol{\theta}' \in \Theta \text{ s.t. } \big(\text{NMI}(\boldsymbol{\theta}') \ge \text{NMI}(\boldsymbol{\theta}) \;\wedge\; \text{Sil}(\boldsymbol{\theta}') \ge \text{Sil}(\boldsymbol{\theta})\big) \\
    &\text{with strict inequality in at least one objective} \Big\}.
  \end{aligned}
  \label{eq:pareto_front}
\end{equation}

\begin{figure}[!htbp]
  \centering
  \begin{subfigure}[b]{0.84\linewidth}
    \includegraphics[width=\linewidth]{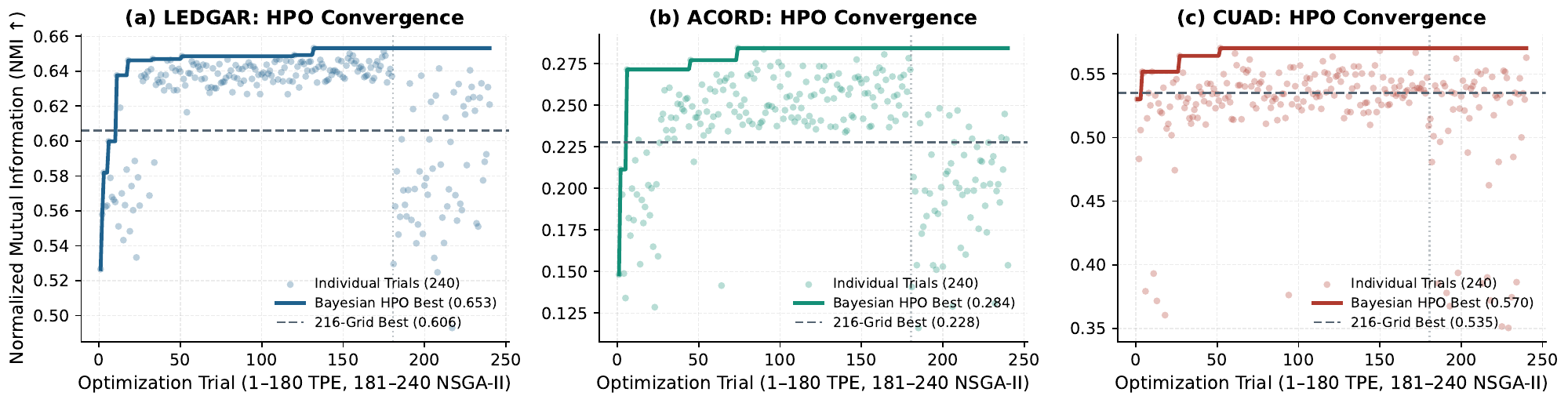}
    \caption{Bayesian TPE (Trials 1--180) and NSGA-II (Trials 181--240) convergence trajectories versus the 216-grid baseline.}
    \label{fig:hpo_history}
  \end{subfigure}
  \vspace{1pt}
  \begin{subfigure}[b]{0.84\linewidth}
    \includegraphics[width=\linewidth]{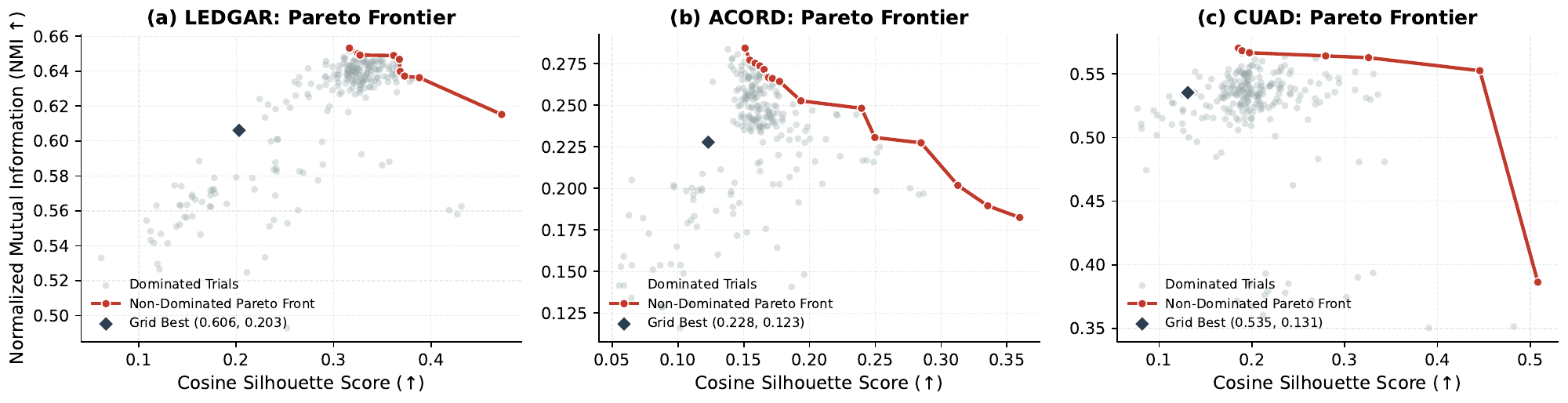}
    \caption{Empirical Multi-Objective Pareto Frontiers over external alignment ($\text{NMI} \uparrow$) and internal compactness ($\text{Silhouette} \uparrow$).}
    \label{fig:hpo_pareto}
  \end{subfigure}
  \vspace{1pt}
  \begin{subfigure}[b]{0.84\linewidth}
    \includegraphics[width=\linewidth]{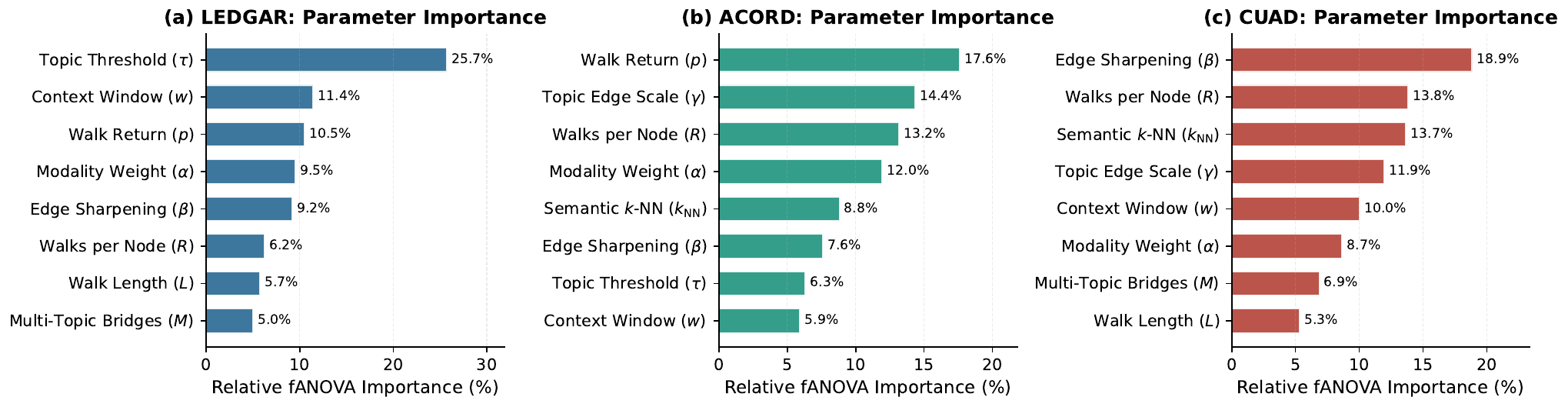}
    \caption{Functional ANOVA (fANOVA) relative hyperparameter importance rankings across LEDGAR, ACORD, and CUAD.}
    \label{fig:hpo_importance}
  \end{subfigure}
  \caption{Comprehensive diagnostics of the 720-trial GPU-accelerated Bayesian TPE and Multi-Objective NSGA-II Pareto optimization across (a--c) LEDGAR ($N=4,000, K=28$), ACORD ($N=2,227, K=7$), and CUAD ($N=407, K=20$).}
  \label{fig:hpo_convergence_pareto}
\end{figure}

Table~\ref{tab:bayesian_hpo_summary} summarizes the complete 720-trial Bayesian and Pareto HPO results, and Figure~\ref{fig:hpo_convergence_pareto} presents the optimization convergence trajectories, empirical non-dominated Pareto frontiers over $(\text{NMI} \uparrow, \text{Cosine Silhouette} \uparrow)$, and functional ANOVA (fANOVA) \citep{hutter2014fanova} hyperparameter importance distributions.

\begin{table}[!htbp]
  \centering
  \caption{Summary of the 720-Trial GPU-Accelerated Bayesian Multivariate TPE and Multi-Objective NSGA-II Pareto Optimization ($240\text{ Trials/Dataset} = 180\text{ TPE} + 60\text{ NSGA-II}$) for TGSF Across LEDGAR, ACORD, and CUAD.}
  \label{tab:bayesian_hpo_summary}
  \resizebox{\linewidth}{!}{%
  \begin{tabular}{@{}llccccc@{}}
    \toprule
    \textbf{Dataset} & \textbf{HPO Statistic / Optimal Regime} & \textbf{NMI ($\uparrow$)} & \textbf{ARI ($\uparrow$)} & \textbf{Silhouette ($\uparrow$)} & \textbf{DBI ($\downarrow$)} & \textbf{CHS ($\uparrow$)} \\
    \midrule
    LEDGAR ($N=4,000, K=28$)
      & 216-Grid Best (MiniLM, $\alpha=0.5$) & $0.606$ & $0.395$ & $0.203$ & $2.421$ & $88.1$ \\
      & Bayesian HPO Best (Trial \#131)      & $\mathbf{0.653}$ & $0.422$ & $0.316$ & $1.916$ & $137.2$ \\
      & Max-ARI Trial (Trial \#50)           & $0.649$ & $\mathbf{0.425}$ & $\mathbf{0.326}$ & $\mathbf{1.894}$ & $\mathbf{141.5}$ \\
      & 240-Trial Mean $\pm$ SD              & $0.618 \pm 0.035$ & $0.372 \pm 0.052$ & $0.293 \pm 0.072$ & $1.988 \pm 0.358$ & $139.4 \pm 38.9$ \\
      & 240-Trial Min -- Max                 & $0.493 \text{ -- } 0.653$ & $0.216 \text{ -- } 0.425$ & $0.061 \text{ -- } 0.472$ & $1.327 \text{ -- } 3.665$ & $44.0 \text{ -- } 433.3$ \\
      & Optimal Hyperparameters ($\boldsymbol{\theta}^*$) & \multicolumn{5}{l}{Legal-BERT, $T=15, k_{\text{NN}}=16, M=5, \tau=0.055, \gamma=0.33, \beta=1.45, d_{\text{graph}}=64, L=15, R=45, p=0.65, q=0.74, w=10, \alpha=0.83$} \\
      & Top fANOVA Importance Drivers        & \multicolumn{5}{l}{Topic Threshold $\tau$ ($25.7\%$), Context Window $w$ ($11.4\%$), Walk Return $p$ ($10.5\%$), Modality Weight $\alpha$ ($9.5\%$), Edge Sharpening $\beta$ ($9.2\%$)} \\
    \midrule
    ACORD ($N=2,227, K=7$)
      & 216-Grid Best (MiniLM, $\alpha=0.5$) & $0.228$ & $0.115$ & $0.123$ & $3.263$ & $89.7$ \\
      & Bayesian HPO Best (Trial \#73)       & $\mathbf{0.284}$ & $0.246$ & $\mathbf{0.151}$ & $\mathbf{2.384}$ & $\mathbf{118.9}$ \\
      & Max-ARI Trial (Trial \#89)           & $0.284$ & $\mathbf{0.260}$ & $0.138$ & $2.451$ & $114.2$ \\
      & 240-Trial Mean $\pm$ SD              & $0.232 \pm 0.036$ & $0.141 \pm 0.057$ & $0.161 \pm 0.043$ & $2.614 \pm 0.368$ & $128.5 \pm 52.9$ \\
      & 240-Trial Min -- Max                 & $0.116 \text{ -- } 0.284$ & $0.010 \text{ -- } 0.260$ & $0.055 \text{ -- } 0.360$ & $1.396 \text{ -- } 3.721$ & $60.0 \text{ -- } 714.4$ \\
      & Optimal Hyperparameters ($\boldsymbol{\theta}^*$) & \multicolumn{5}{l}{Legal-BERT, $T=25, k_{\text{NN}}=6, M=1, \tau=0.058, \gamma=2.56, \beta=0.88, d_{\text{graph}}=256, L=45, R=30, p=0.40, q=0.95, w=4, \alpha=0.19$} \\
      & Top fANOVA Importance Drivers        & \multicolumn{5}{l}{Walk Return $p$ ($17.6\%$), Topic Edge Scale $\gamma$ ($14.4\%$), Walks/Node $R$ ($13.2\%$), Modality Weight $\alpha$ ($12.0\%$), $k_{\text{NN}}$ ($8.8\%$)} \\
    \midrule
    CUAD ($N=407, K=20$)
      & 216-Grid Best (MiniLM, $\alpha=0.5$) & $0.535$ & $0.253$ & $0.131$ & $2.667$ & $9.8$ \\
      & Bayesian HPO Best (Trial \#52)       & $\mathbf{0.570}$ & $0.277$ & $0.185$ & $2.306$ & $15.6$ \\
      & Max-ARI Trial (Trial \#172)          & $0.567$ & $\mathbf{0.282}$ & $\mathbf{0.197}$ & $\mathbf{2.241}$ & $\mathbf{16.4}$ \\
      & 240-Trial Mean $\pm$ SD              & $0.523 \pm 0.043$ & $0.234 \pm 0.038$ & $0.206 \pm 0.058$ & $2.207 \pm 0.309$ & $18.4 \pm 8.4$ \\
      & 240-Trial Min -- Max                 & $0.351 \text{ -- } 0.570$ & $0.084 \text{ -- } 0.282$ & $0.077 \text{ -- } 0.508$ & $1.096 \text{ -- } 3.260$ & $5.8 \text{ -- } 76.7$ \\
      & Optimal Hyperparameters ($\boldsymbol{\theta}^*$) & \multicolumn{5}{l}{all-mpnet-base-v2, $T=15, k_{\text{NN}}=11, M=3, \tau=0.113, \gamma=0.35, \beta=1.98, d_{\text{graph}}=128, L=45, R=35, p=3.09, q=0.45, w=3, \alpha=0.77$} \\
      & Top fANOVA Importance Drivers        & \multicolumn{5}{l}{Edge Sharpening $\beta$ ($18.9\%$), Walks/Node $R$ ($13.8\%$), $k_{\text{NN}}$ ($13.7\%$), Topic Edge Scale $\gamma$ ($11.9\%$), Context Window $w$ ($10.0\%$)} \\
    \bottomrule
  \end{tabular}%
  }
\end{table}

Table~\ref{tab:bayesian_hpo_summary} and Figure~\ref{fig:hpo_convergence_pareto} yield three key methodological insights:
\begin{enumerate}
  \item \textbf{Substantial Gains Over Exhaustive Fixed Grids:} Across all three legal benchmarks, Multivariate TPE rapidly surpasses the 216-configuration grid search within the first $10\text{--}30$ trials and continues to refine the joint manifold up to Trial $180$ (Figure~\ref{fig:hpo_history}). On LEDGAR, NMI improves from $0.606 \to 0.653$ ($+7.8\%$), ARI from $0.395 \to 0.425$, and Silhouette from $0.203 \to 0.316$; on ACORD, NMI improves from $0.228 \to 0.284$ ($+24.8\%$) and ARI more than doubles from $0.115 \to 0.260$ ($+126.1\%$); and on CUAD, NMI improves from $0.535 \to 0.570$ ($+6.5\%$) and ARI from $0.253 \to 0.282$.
  \item \textbf{Pareto Dominance Over Fixed-Grid Baselines:} As shown in Figure~\ref{fig:hpo_pareto}, the 216-grid best configurations (dark diamonds) lie strictly inside the dominated interior of the $(\text{NMI}, \text{Silhouette})$ objective space across all three datasets. The non-dominated Pareto frontiers (crimson curves) reveal that practitioners can select operating points that simultaneously achieve higher NMI and substantially tighter cluster geometry (reaching Silhouette scores up to $0.472$ on LEDGAR, $0.360$ on ACORD, and $0.508$ on CUAD).
  \item \textbf{Domain-Specific Encoder Specialization and fANOVA Drivers:} Bayesian HPO automatically selects domain-pretrained Legal-BERT for short, specialized legal clauses and provisions (LEDGAR and ACORD), while selecting general-purpose high-capacity all-mpnet-base-v2 for full multi-page commercial agreements (CUAD). Furthermore, fANOVA importance analysis (Figure~\ref{fig:hpo_importance}) confirms that continuous graph-structural and fusion parameters (specifically the topic threshold $\tau$ at $25.7\%$ on LEDGAR, second-order return bias $p$ at $17.6\%$ on ACORD, edge temperature sharpening $\beta$ at $18.9\%$ on CUAD, topic edge scale $\gamma$ at $11.9\text{--}14.4\%$, and modality fusion weight $\alpha$ at $8.7\text{--}12.0\%$) account for the majority of variance in clustering performance.
\end{enumerate}

\subsection{Ablation Study: Isolating Component Contributions}
\label{subsec:ablation}

To evaluate the contribution of each architectural component of TGSF, we conducted an ablation study across all three datasets comparing: (1) the lexical baseline (TF-IDF + KMeans), (2) the semantic-only model (Top2Vec + KMeans, corresponding to $\alpha = 0$ in Equation~\eqref{eq:concat}), (3) the structural-only model (Node2Vec + KMeans on $G$, corresponding to $\alpha = 1$), (4) balanced spherical fusion (TGSF Grid, $\alpha = 0.5$), and (5) Bayesian-optimized spherical fusion (TGSF Bayesian HPO). Table~\ref{tab:ablation_study} reports the resulting internal and external clustering metrics.

\begin{table}[!htbp]
\centering
\caption{Ablation Study Isolating Lexical, Semantic-Only ($\alpha=0$), Structural-Only ($\alpha=1$), Balanced Spherical Fusion ($\alpha=0.5$), and Bayesian HPO Components of TGSF Across LEDGAR, ACORD, and CUAD.}
\label{tab:ablation_study}
\resizebox{\linewidth}{!}{%
\begin{tabular}{@{}llrrrrr@{}}
\toprule
\textbf{Dataset} & \textbf{Model Configuration} & \multicolumn{1}{c}{\textbf{Silhouette ($\uparrow$)}} & \multicolumn{1}{c}{\textbf{CHS ($\uparrow$)}} & \multicolumn{1}{c}{\textbf{DBI ($\downarrow$)}} & \multicolumn{1}{c}{\textbf{NMI ($\uparrow$)}} & \multicolumn{1}{c}{\textbf{ARI ($\uparrow$)}} \\ \midrule
LEDGAR ($N=4,000, K=28$)
  & TF-IDF + KMeans (Lexical Baseline)        & 0.085 &   25.9 & 4.564 & 0.523 & 0.213 \\
  & Top2Vec + KMeans (Semantic-Only, $\alpha=0$) & 0.117 &   61.2 & 3.063 & 0.553 & 0.324 \\
  & Node2Vec + KMeans (Structural-Only, $\alpha=1$) & 0.271 &  110.5 & 2.015 & 0.593 & 0.356 \\
  & TGSF (Balanced Spherical Fusion, $\alpha=0.5$) & 0.203 &   88.1 & 2.421 & 0.606 & 0.395 \\
  & TGSF (Bayesian HPO Spherical Fusion, $\alpha=0.83$) & \textbf{0.316} & \textbf{137.2} & \textbf{1.916} & \textbf{0.653} & \textbf{0.422} \\ \midrule
ACORD ($N=2,227, K=7$)
  & TF-IDF + KMeans (Lexical Baseline)        & 0.045 &   26.6 & 5.084 & 0.173 & 0.145 \\
  & Top2Vec + KMeans (Semantic-Only, $\alpha=0$) & 0.095 &   76.5 & 3.480 & 0.200 & 0.081 \\
  & Node2Vec + KMeans (Structural-Only, $\alpha=1$) & \textbf{0.188} & \textbf{119.3} & 2.668 & 0.185 & 0.073 \\
  & TGSF (Balanced Spherical Fusion, $\alpha=0.5$) & 0.123 &   89.7 & 3.263 & 0.228 & 0.115 \\
  & TGSF (Bayesian HPO Spherical Fusion, $\alpha=0.19$) & 0.151 &  118.9 & \textbf{2.384} & \textbf{0.284} & \textbf{0.246} \\ \midrule
CUAD ($N=407, K=20$)
  & TF-IDF + KMeans (Lexical Baseline)        & 0.046 &    2.7 & 4.757 & 0.413 & 0.145 \\
  & Top2Vec + KMeans (Semantic-Only, $\alpha=0$) & 0.052 &    5.2 & 3.469 & 0.483 & 0.201 \\
  & Node2Vec + KMeans (Structural-Only, $\alpha=1$) & \textbf{0.208} &   15.5 & \textbf{2.098} & 0.514 & 0.221 \\
  & TGSF (Balanced Spherical Fusion, $\alpha=0.5$) & 0.131 &    9.8 & 2.667 & 0.535 & 0.253 \\
  & TGSF (Bayesian HPO Spherical Fusion, $\alpha=0.77$) & 0.185 & \textbf{15.6} & 2.306 & \textbf{0.570} & \textbf{0.277} \\ \bottomrule
\end{tabular}%
}
\end{table}

The ablation results in Table~\ref{tab:ablation_study} highlight three consistent findings:
\begin{enumerate}
  \item \textbf{Dense Semantic Embeddings Overcome Lexical Sparsity:} Replacing sparse TF-IDF vectors with dense Top2Vec embeddings ($\alpha=0$) improves both internal geometry and external label alignment across all datasets (raising NMI from $0.523$ to $0.553$ and ARI from $0.213$ to $0.324$ on LEDGAR, and raising NMI from $0.413$ to $0.483$ on CUAD).
  \item \textbf{Graph Propagation Sharpens Cluster Structure:} Applying Node2Vec over the multi-relational graph $G$ ($\alpha=1$) acts as a relational smoothing operator, increasing geometric compactness (raising Silhouette from $0.117$ to $0.271$ on LEDGAR and from $0.052$ to $0.208$ on CUAD) while propagating community signals across $k$-NN neighborhoods.
  \item \textbf{Spherical Fusion Outperforms Unimodal Representations:} Combining both modalities on the joint unit hypersphere outperforms both semantic-only and structural-only ablations on external validation metrics across all three datasets under balanced weighting ($\alpha=0.5$), and improves further under Bayesian HPO calibration: on LEDGAR, NMI reaches $0.653$ ($+10.0$ points over semantic-only and $+6.0$ points over structural-only) and ARI reaches $0.422$ (up to $0.425$); on ACORD ($K=7$), NMI reaches $0.284$ ($+8.4$ points over semantic-only and $+9.9$ points over structural-only) and ARI reaches $0.246$ (up to $0.260$); and on CUAD ($K=20$), NMI reaches $0.570$ ($+8.7$ points over semantic-only and $+5.6$ points over structural-only) and ARI reaches $0.277$ (up to $0.282$).
\end{enumerate}

\section{Discussion and Practical Guidelines}
\label{sec:discussion}

Our empirical findings offer several actionable insights and practical guidelines for data mining researchers and legal NLP practitioners:
\begin{itemize}
  \item \textbf{Encoder Specialization Across Legal Document Lengths:} As demonstrated in Table~\ref{tab:bayesian_hpo_summary}, domain-pretrained Legal-BERT \citep{chalkidis2020legalbert} yields the strongest representations on short clause- and provision-level benchmarks (LEDGAR and ACORD), where specialized statutory and contractual terminology dominates within a single $512$-token window. Conversely, on full-length commercial agreements in CUAD, general-purpose all-mpnet-base-v2 outperforms Legal-BERT, reflecting its stronger long-range discourse aggregation across multi-page contract contexts; integrating long-context legal encoders (e.g., $4,096$-token transformer backbones) is a promising avenue to unify clause- and contract-level performance.
  \item \textbf{Reconciling Internal Compactness and External Alignment via Pareto Frontiers:} A recurring pitfall in unsupervised clustering is relying solely on internal Silhouette or CHS scores to compare representations of different dimensionality or topological connectivity. As shown in Table~\ref{tab:multi_dataset_benchmark} and Figure~\ref{fig:hpo_pareto}, mapping the multi-objective $(\text{NMI}, \text{Silhouette})$ Pareto frontier enables practitioners to select knee-point configurations that simultaneously preserve high external category alignment and strong geometric cluster separation.
  \item \textbf{Computational Complexity and Scalability:} Constructing the semantic $k$-NN graph requires $O(N^2 d_{\text{sem}})$ exact inner products (or $O(N \log N)$ via approximate nearest-neighbor indexing such as FAISS or HNSW), while Node2Vec random-walk simulation and Skip-gram training scale linearly with corpus size as $O(R L w (N + T))$. Combined with representational saturation at $70\text{--}80\%$ corpus volume (Figure~\ref{fig:DB_S_metrics}), embedding and clustering $N=4,000$ LEDGAR provisions completes in under $18$ seconds per configuration on a single GPU workstation.
  \item \textbf{Human-in-the-Loop Deployment and Overlapping Communities:} From a legal informatics perspective, TGSF addresses both scalability without annotation and interpretability by exposing an explicit intermediate graph structure comprising the multi-relational document--topic network (Figure~\ref{fig:bipartite}) and category-annotated topic hubs (Figure~\ref{fig:cuadlabels}). Legal practitioners can inspect top keywords and prototype clauses of each discovered topic hub, examine why two contracts are linked via shared topic hubs or semantic $k$-NN edges, and merge or split clusters interactively during e-discovery and contract review. Because hard KMeans partitioning assigns each document to a single primary cluster, clauses spanning multiple legal obligations (e.g., a provision combining Indemnification and Limitation of Liability) cannot be assigned to multiple categories simultaneously; replacing hard partitioning with fuzzy $c$-means or overlapping community detection on $G$ is a natural extension.
\end{itemize}

\section{Conclusion}
\label{sec:conclusion}

We have presented and evaluated Topic-Graph Spherical Fusion (TGSF), a dual-manifold unsupervised representation learning framework that unifies dense topic-semantic embeddings from Top2Vec with second-order biased random-walk structural embeddings from Node2Vec on a joint unit hypersphere for unsupervised legal document clustering. By constructing a connected multi-relational graph $G = (V_D \cup V_T, E_{DT} \cup E_{DD}, W)$ ($\dim\ker(\mathcal{L}) = 1$) that integrates multi-topic bridges with temperature-sharpened semantic $k$-Nearest Neighbor ($k$-NN) document--document edges, and fusing both $L_2$-normalized modalities via chordal-distance kernel decomposition on $\mathbb{S}^{d_{\text{sem}}+d_{\text{graph}}-1}$, TGSF enables random walks and spherical partitioning to jointly capture macro-level thematic communities and micro-level semantic proximity. Evaluations across the LEDGAR, ACORD, and CUAD legal benchmarks demonstrate that TGSF generates coherent, interpretable topic clusters that align substantially more closely with ground-truth legal categories than lexical, probabilistic, matrix-factorization, or unimodal baselines. Supported by representational saturation at $70\text{--}80\%$ corpus volume and systematic Bayesian and multi-objective Pareto optimization, TGSF provides a scalable, reproducible methodology for unsupervised knowledge discovery in legal text mining.

\section*{Data and Code Availability}

All benchmark datasets analyzed in this study are publicly available: LEDGAR via the LexGLUE benchmark (\url{https://huggingface.co/datasets/coastalcph/lex_glue}), ACORD via the Atticus Project (\url{https://huggingface.co/datasets/theatticusproject/acord}), and CUAD via the Atticus Project (\url{https://www.atticusprojectai.org/cuad}). The complete reproducible CPU/GPU source code, hyperparameter optimization scripts, and plotting pipelines are publicly available at \url{https://github.com/deepbas/TGSF-Legal-Clustering}.

\appendix

\section{Theoretical Foundations of UMAP}
\label{sec:UMAP-theory}

Uniform Manifold Approximation and Projection (UMAP) \citep{mcinnes2018umap} is a key tool in our pipeline for transforming high-dimensional document representations into an intelligible low-dimensional visualization that preserves the data's inherent topological structure. Grounded in Riemannian geometry and algebraic topology, UMAP assumes that observed high-dimensional data points lie on an underlying manifold of lower intrinsic dimension and learns a low-dimensional embedding $\mathbf{Y} = \{\mathbf{y}_i\}_{i=1}^N \subset \mathbb{R}^d$ ($d \in \{2, 3\}$) that preserves its fuzzy simplicial set structure.

Statistically, UMAP's procedure involves two main stages. First, it constructs a weighted fuzzy topological graph representation of the data in the high-dimensional space. For each data point $\mathbf{h}_i$, UMAP computes a directed conditional probability $p_{j\mid i}$ representing its similarity to $\mathbf{h}_j$ via a locally adapted exponential kernel:
\begin{equation}
  p_{j\mid i} = \exp\!\left(-\frac{\max(0, d(\mathbf{h}_i, \mathbf{h}_j) - \rho_i)}{\sigma_i}\right), \quad
  \rho_i = \min_{j \in \mathcal{N}(\mathbf{h}_i)} d(\mathbf{h}_i, \mathbf{h}_j), \quad
  \sum_{j \in \mathcal{N}(\mathbf{h}_i)} p_{j\mid i} = \log_2(n_{\text{neighbors}}).
  \label{eq:umap_pji}
\end{equation}
Here, $d(\mathbf{h}_i, \mathbf{h}_j)$ is the pairwise distance (cosine distance in our pipeline), $\rho_i$ is the distance from $\mathbf{h}_i$ to its nearest neighbor (ensuring local connectivity), and $\sigma_i$ is a point-specific bandwidth chosen to normalize distances based on local data density. To make the graph undirected, these conditional probabilities are symmetrized via the fuzzy set union (probabilistic t-conorm):
\begin{equation}
  p_{ij} = p_{j\mid i} + p_{i\mid j} - p_{j\mid i}p_{i\mid j}.
  \label{eq:umap_sym}
\end{equation}

Second, UMAP defines an analogous similarity measure $q_{ij}$ for the corresponding points $(\mathbf{y}_i, \mathbf{y}_j)$ in the target low-dimensional embedding space ($\mathbb{R}^2$ or $\mathbb{R}^3$) using a differentiable family of curves closely related to the Student's $t$-distribution:
\begin{equation}
  q_{ij} = \left(1 + a \|\mathbf{y}_i - \mathbf{y}_j\|_2^{2b}\right)^{-1},
  \label{eq:umap_qij}
\end{equation}
where $a$ and $b$ are hyperparameters fitted to approximate an indicator function with minimum-distance margin $d_{\min}$, controlling the tight packing of points in the low-dimensional space. UMAP then minimizes the fuzzy set cross-entropy $\mathcal{L}_{\text{UMAP}}(\mathbf{Y})$ between the high-dimensional topological weights $p_{ij}$ and the low-dimensional weights $q_{ij}$ via stochastic gradient descent:
\begin{equation}
  \mathcal{L}_{\text{UMAP}}(\mathbf{Y}) = \sum_{i \neq j} \left[ p_{ij} \log\!\left(\frac{p_{ij}}{q_{ij}}\right) + (1 - p_{ij}) \log\!\left(\frac{1 - p_{ij}}{1 - q_{ij}}\right) \right].
  \label{eq:umap_loss}
\end{equation}
This objective penalizes topological mismatches: the first term applies an attractive force between points that are proximate in the high-dimensional space (high $p_{ij}$), while the second term applies a repulsive force between points that are distant (low $p_{ij}$), preserving both local neighborhood cohesion and global inter-cluster separation.

\section{Reproducibility and Hyperparameter Specifications}
\label{sec:hyperparams}

For full reproducibility, this appendix summarizes the hyperparameter configurations and software environment used across all experiments:
\begin{itemize}
  \item \textbf{Top2Vec and Semantic Backbones:} We used the all-MiniLM-L6-v2 sentence-transformer embedding model ($d_{\text{sem}} = 384$) \citep{reimers2019sentencebert} for the 216-configuration factorial grid search, and expanded the 720-trial Bayesian HPO search space to include all-mpnet-base-v2 ($d_{\text{sem}} = 768$) and Legal-BERT (nlpaueb/legal-bert-base-uncased, $d_{\text{sem}} = 768$) \citep{chalkidis2020legalbert} with a minimum word count of $5$.
  \item \textbf{Multi-Relational Graph Construction ($G$):} In the multi-relational graph $G = (V_D \cup V_T, E_{DT} \cup E_{DD}, W)$ (Equations~\eqref{eq:edt_weights}--\eqref{eq:edd_weights}), we evaluated top-$M$ document--topic bridges with $M \in \{1, 2, 3\}$ in the factorial grid ($M \in \{1, \dots, 5\}$ in Bayesian HPO, for cosine similarity $> \tau$ with $\tau \in [0.02, 0.25]$ and bridge scale $\gamma \in [0.3, 3.5]$) and cosine-weighted semantic $k$-NN document--document edges with $k_{\text{NN}} \in \{2, 5, 10, 15\}$ in the grid ($k_{\text{NN}} \in [2, 22]$ and temperature sharpening $\beta \in [0.5, 2.5]$ in Bayesian HPO).
  \item \textbf{Node2Vec Biased Random Walks:} Across the factorial grid search, graph embeddings were evaluated over dimensionalities $d_{\text{graph}} \in \{32, 64, 128\}$, random-walk lengths $L \in \{15, 30\}$, and $R \in \{10, 20, 40\}$ walks per node, using a Skip-gram context window size $w = 5$ and $p = q = 1.0$. In the Bayesian Multivariate TPE and NSGA-II Pareto studies (Equation~\eqref{eq:hpo_pareto}), $d_{\text{graph}} \in \{32, 64, 128, 256\}$, $(p, q) \in [0.25, 4.0]^2$, $L \in [10, 45]$, $R \in [10, 45]$, $w \in [3, 10]$, and $\alpha \in [0.10, 0.90]$ were jointly optimized using Optuna 5.0 \citep{akiba2019optuna}.
  \item \textbf{Spherical Fusion, KMeans, and UMAP:} Modality weight $\alpha$ was fixed at $0.5$ in the factorial grid and tuned over $\alpha \in [0.10, 0.90]$ in Bayesian HPO. KMeans (Equation~\eqref{eq:kmeans_obj}) was initialized using k-means++ with $10\text{--}15$ restarts (random seed $42$). UMAP projections used $n_{\text{neighbors}} = 15$, minimum distance $d_{\min} = 0.1$, and cosine metric.
  \item \textbf{Hardware and Software Stack:} All experiments were conducted in a Python 3.10 environment accelerated on an NVIDIA GB10 Grace Blackwell GPU (CUDA 13.0) using Top2Vec 1.0.29, NetworkX 3.x, node2vec 0.4.1, gensim 4.x, scikit-learn 1.x, sentence-transformers 2.x, umap-learn 0.5.x, PyTorch 2.14, and Optuna 5.0.
\end{itemize}


\end{document}